\documentclass[a4paper,UKenglish,cleveref, autoref, thm-restate]{tgdk-v2021}

\usepackage{subfiles,dcolumn,xcolor,tcolorbox,multirow,graphicx}
\usepackage{tabularx}
\usepackage{booktabs}
\usepackage{makecell}
\definecolor{lightlavender}{HTML}{D4C4FF}  
\usepackage[utf8]{inputenc}
\usepackage{xcolor}
\usepackage{listings}

\lstdefinelanguage{CSV}{%
  moredelim=[l][\color{blue}]{,},%
  morestring=[b]",%
  stringstyle=\color{orange},%
  columns=fullflexible,%
  keepspaces=true,%
}[keywords,comments,strings]

\lstdefinelanguage{XML}{
  morestring=[b]",
  morestring=[s]{>}{<},
  morecomment=[s]{<?}{?>},
  commentstyle=\color{gray}\upshape,
  stringstyle=\color{orange},
  identifierstyle=\color{black},
  moredelim=[s][\color{blue}]{<}{>},
  moredelim=[s][\color{blue}]{</}{>},
  literate=
    {á}{{\'a}}1 {é}{{\'e}}1 {í}{{\'i}}1 {ó}{{\'o}}1 {ú}{{\'u}}1
    {Á}{{\'A}}1 {É}{{\'E}}1 {Í}{{\'I}}1 {Ó}{{\'O}}1 {Ú}{{\'U}}1
    {ñ}{{\~n}}1 {Ñ}{{\~N}}1 {ü}{{\"u}}1,
}

\newtcolorbox{consultabox}[1]{
    colback=white,
    colframe=lightlavender,
    title={\textbf{Prompt template}},
    coltitle=black,
    fonttitle=\bfseries,
    boxrule=0.5pt,
    arc=3pt, 
    width=\textwidth, 
    float,
    left=5pt,
    right=5pt,
    top=5pt,
    bottom=5pt,
    before upper={\parindent0pt},
    fontupper=\itshape
}

\title{BLINKG: A Benchmark for LLM-Integrated Knowledge Graph Generation}

\titlerunning{BLINKG: A Benchmark for LLM-Integrated KG Generation}

\author{Carla Castedo}{Centro Singular de Investigación en Tecnoloxías Intelixentes (CiTIUS), Universidade de Santiago de Compostela, Spain}{carlacastedo.pereira@usc.es}{
https://orcid.org/0009-0009-6158-1068}{}

\author{Enrique Iglesias}{L3S Research Center Germany, Hannover, Germany}{iglesias@l3s.de}{https://orcid.org/0000-0002-8734-3123}{}

\author{Manuel Lama}{Departamento de Electrónica e Computación, Universidade de Santiago de Compostela, Spain \and Centro Singular de Investigación en Tecnoloxías Intelixentes (CiTIUS), Universidade de Santiago de Compostela, Spain}{manuel.lama@usc.es}{https://orcid.org/0000-0001-7195-6155}{}

\author{Alberto Bugarín-Diz}{Departamento de Electrónica e Computación, Universidade de Santiago de Compostela, Spain \and Centro Singular de Investigación en Tecnoloxías Intelixentes (CiTIUS), Universidade de Santiago de Compostela, Spain}{alberto.bugarin.diz@usc.es}{https://orcid.org/0000-0003-3574-3843}{}

\author{Maria-Esther Vidal}{L3S Research Center Germany, Hannover, Germany \and TIB Leibniz Information Centre for Science and Technology, Hannover, Germany \and Leibniz University, Hannover, Germany}{maria.vidal@tib.eu}{https://orcid.org/0000-0003-1160-8727}{}

\author{David Chaves-Fraga}{Departamento de Electrónica e Computación, Universidade de Santiago de Compostela, Spain \and Centro Singular de Investigación en Tecnoloxías Intelixentes (CiTIUS), Universidade de Santiago de Compostela, Spain}{david.chaves@usc.es}{https://orcid.org/0000-0003-3236-2789}{}

\authorrunning{C. Castedo, E. Iglesias, M. Lama, A. Bugarín-Diz, M.E. Vidal and D. Chaves-Fraga} 

\Copyright{Carla Castedo, Alberto Bugarín-Diz, Manuel Lama, Enrique Iglesias, Maria-Esther Vidal and David Chaves-Fraga} 


\category{} 

\relatedversion{} 

\supplement{All the resources used in this paper are openly available under Apache 2.0 at \url{https://github.com/citiususc/blinkg}}
\supplementdetails[linktext={Persistent Benchmark URI}]{Dataset}{https://doi.org/10.5281/zenodo.15971734} 

\funding{The members from University of Santiago de Compostela are funded by the Agencia Estatal de Investigaci\'on (Spain) (PID2023-149549NB-I00 \& CPP2024-011786), the Xunta de Galicia — Conseller\'{\i}a de Cultura, Educaci\'on, Formaci\'on Profesional e Universidades (Centro de investigaci\'on de Galicia accreditation 2024–2027 ED431G-2023/04 and Reference Competitive Group accreditation 2022–2026, ED431C 2022/19) and the European Union (European Regional Development Fund — ERDF).}

\acknowledgements{We want to thank Jhon Toledo, Ana Iglesias-Molina, Oscar Corcho and Daniel Garijo for the discussions and ideas behind the proposed benchmark. We also want to thank Javier Garea for his support and feedback in the use of BLINKG}

\nolinenumbers 

\Volume{}
\Issue{?}
\Article{?}
\DateSubmission{July 2025}
\DateAcceptance{18 March 2026}

\begin{document}

\maketitle

\begin{abstract}
Generating Knowledge Graphs (KGs) remains one of the most time-consuming and labor-intensive tasks for knowledge engineers, as they need to identify semantic equivalences between input data sources and ontology terms. While declarative solutions (e.g., RML, SPARQL-Anything) have helped to generalize this process, aligning input schema elements with ontology terms still involves intricate transformations and requires considerable manual effort. With the advent of Large Language Models (LLMs), there is growing interest in leveraging their capabilities to assist KG engineers. Although some studies have explored using LLMs to automate KG construction, there is still no standardized framework for assessing how effectively they establish correspondences between data schemes and ontology concepts. Therefore, in this paper, we propose BLINKG, a benchmark designed to evaluate the mapping capabilities of LLMs in constructing KGs from heterogeneous data sources. The benchmark includes a set of scenarios with increasing complexity, based on real-world use cases. We conduct an extensive experimental evaluation of several state-of-the-art LLMs using BLINK and observe that they already offer promising solutions. However, their performance remains limited in complex scenarios. Thanks to this benchmark we can already asses the current capabilities of LLMs for KG construction. Additionally, we define a set of requirements for achieving (semi)automated (LLM-driven) KG construction, opening new research lines in this area.
\end{abstract}

\section{Introduction}
\label{sec:introduction}



Knowledge engineering (KE) covers tasks required to make data and knowledge computationally accessible, tasks that traditionally depend on extensive manual work by domain experts and knowledge engineers. Large Language Models (LLMs) are increasingly used to support these processes, such as formulating competency questions~\cite{rebboud2024can}, taking on expert roles in ontology engineering methodologies~\cite{fathallah2024neon,zhang2024ontochat}, and verbalizing knowledge graph queries~\cite{perevalov2024understanding}. The growing number of LLM-based KE approaches has also motivated new benchmarking and evaluation initiatives like Alharbi et al.~\cite{Alharbi2024Characteristics}, Garijo et al.~ \cite{garijo2024llms}, Herwanto et al.~\cite{Herwanto2024Ontology}, Rebboud et al.~\cite{Rebboud2024Benchmarking}, and Tsaneva et al.~\cite{Tsaneva2024Benchmarking} to assess their current capabilities and limitations.

Constructing KG from heterogeneous data sources is one of the most time-consuming and manual tasks that a knowledge engineer must perform~\cite{dimou2022declarative}. The difficulty of the task lies in the need to understand the ontology terms and map the input data to these terms~\cite{poggi2008linking}. In some cases, the process can be relatively simple (e.g., mapping a table Sport to an ontology class Sport), but in many situations it requires a deeper understanding of the data and the domain. Often, domain experts are needed to correctly interpret implicit semantics and resolve ambiguities between the data and the ontology. In large-scale, real-world projects~\cite{chaves2022systematic,rojas2021leveraging}, this task is often a bottleneck, with the mapping phase frequently exceeding six person-months~\cite{chaves2022systematic}.

Several solutions have recently explored using LLMs to reduce the effort required for semantic alignment between data sources and ontologies. For example, in Hofer et al.~\cite{hofer2024towards}, the authors propose an automated KG construction pipeline where RML mappings~\cite{iglesias2023rml} are generated by an LLM using the ontology and source data as inputs. In Schmidt et al.~\cite{schmidt2025llm}, the generation of YARRRML mappings~\cite{heyvaert2018declarative} with LLMs is explored in the manufacturing domain. Similarly, [R2]RML-ChatGPT~\cite{randles2024r2} presents a framework to refine mapping rules using ChatGPT. Recently, Freund et al.~\cite{freund2025mapping} evaluates its proposal (the \textit{ReMap} tool) against LLM-based RML~\cite{iglesias2023rml} mapping generation. However, all these solutions have been evaluated across diverse scenarios and use cases, using different parameters and metrics, which makes it difficult to compare them fairly. In the SW and KG community, SemTab\footnote{\url{https://github.com/sem-tab-challenge}} is positioned as an initiative and challenge for automatic annotation of tabular data, providing its own benchmark~\cite{jimenez2020semtab} for testing solutions that automate table interpretation. However, it typically targets an already created knowledge graph (e.g., Wikidata) rather than a non-populated ontology, focuses on three specific tasks (Cell Entity Annotation, Cell Property Annotation, and Cell Type Annotation), and most of the approaches are implemented using end-to-end pipelines, limiting traceability and transparency.

In this context, to assess the potential of LLMs for generating semantic mappings from heterogeneous data sources to ontology terms, we present BLINKG, a \textbf{B}enchmark for \textbf{L}LM-\textbf{In}tegrated \textbf{K}nowledge \textbf{G}raph Generation. The goal is to support the generation of explicit and declarative mappings, providing traceability and transparency in the knowledge graph construction process, as opposed to black-box end-to-end approaches. The main contributions of this work are summarized as follows:
\begin{itemize}
    \item \textbf{C1:} A comprehensive and domain-agnostic framework composed of multiple scenarios, gold standards, and evaluation metrics designed to assess the behavior of LLMs in KG construction
    \item \textbf{C2:} Three progressively complex scenarios that reflect real-world challenges in KG construction, each covering a representative set of typical mapping tasks.
    \item \textbf{C3:} Open and reusable resources, fostering reproducibility and facilitating adoption and extension by the community\footnote{\url{https://github.com/citiususc/blinkg}}.
    \item \textbf{C4:} A comprehensive evaluation of six state-of-the-art LLMs, analyzing their performance and generalization capabilities, focusing on traceability and transparency.
    \item \textbf{C5:} Practical recommendations, lessons learned, and new research directions for advancing the (semi-)automation of LLM-driven knowledge graph construction.
\end{itemize}

The paper is structured as follows: Section 2 describes the related work on the automation of KG Construction, with special focus on novel proposals that use LLMs. Section 3 describes the BLINKG benchmark, with the proposed scenarios and metrics. Section 4 presents the experimental evaluation of our benchmark over state-of-the-art LLMs, and Section 5 reports the main points of discussion. We finalize the paper with the conclusions and future work in Section 6. 

\section{Related Work}
\label{sec:related-work}

Despite recent advances in automatic knowledge extraction, the creation of knowledge graphs remains an inherently manual and resource-intensive process. The main challenge lies in establishing semantic correspondences between heterogeneous data sources and the concepts defined in the target ontologies. In complex scenarios, this process requires not only a knowledge engineer but also a domain expert who can accurately identify and validate these mappings. Although declarative tools help streamline and simplify the process~\cite{VANASSCHE2023}, it remains a challenging task that heavily relies on domain expertise. Existing work can be roughly grouped into (i) approaches that focus on mapping generation, mostly focused on relational databases as input sources, (ii) benchmarks and datasets that evaluate partial aspects of the KG Construction process, and (iii) more recent LLM-based systems, which exhibit different strengths and failure modes from traditional methods.

With the introduction of R2RML~\cite{R2RML} and Direct Mapping~\cite{arenas2012direct}, W3C recommendations for defining mappings between relational databases (RDB) and ontologies in the context of Ontology-Based Data Access (OBDA)~\cite{xiao2018ontology}, a number of approaches emerged aiming to automate the creation of these mapping rules. Most of these approaches followed the directives of Direct Mapping, which defines a set of rules to produce a plain RDF representation of a relational database without considering any ontology. MIRROR~\cite{de2015mirror}, D2RQ~\cite{bizer2004d2rq}, and Ontop~\cite{calvanese2016ontop} follow a similar approach, extracting from the Relational Database (RDB) schema a target ontology and the mapping correspondences.
On the other hand, AutoMap4OBDA~\cite{sicilia2016automap4obda} and BootOX~\cite{jimenez2015bootox} consider an input ontology and generate actual R2RML mappings from the RDB.
However, these solutions are limited to relational databases and rely heavily on heuristic-based approaches for mapping generation, which significantly constrains their applicability and flexibility in broader Knowledge Graph Construction (KGC) workflows that involve multiple formats, complex transformations, and non-relational sources.

All these proposals used RODI for the evaluation, a benchmark for RDB-to-ontology mapping generation~\cite{pinkel2017rodi}. RODI is designed to evaluate the quality of system-generated relational-to-ontology mappings, aiming to provide a generic and comparable evaluation framework for mapping generation systems. RODI employs an end-to-end evaluation approach, assessing mapping utility by comparing the accuracy of SPARQL query results over the generated RDF data against reference SQL query results on the original relational database. Similar to our proposal the benchmark incorporates diverse test scenarios from domains such as scientific conferences, geographical data, and oil and gas exploration. These scenarios are configured with databases, ontologies, and specific query workloads designed to test a variety of mapping challenges, including naming conflicts, structural RDB heterogeneity (e.g., normalization, denormalization, class hierarchies, key conflicts, and dependency conflicts). However, the authors acknowledge certain limitations in the scope of evaluation, particularly concerning the complexity of data transformations. RODI explicitly excluded complex data transformations such as unit conversions, string cleaning, or string compositions from its default benchmark configurations. This exclusion was justified by the observation that, at the time of its publication, ``no current relational-to-ontology mapping generation system implements any such transformation functionality''. Therefore, the benchmark primarily focused on evaluating systems' capabilities in RDB-schema-level matching and mapping rule generation, often resolvable through heuristic techniques, rather than addressing more complex data-level transformations or advanced relationships beyond basic structural mappings. In summary, RODI focuses on relational-to-ontology mappings, primarily evaluating schema-level correspondences and basic structural mappings over relational databases. BLINKG shares this emphasis on mapping generation, but generalises it beyond RDBs to multiple data formats (CSV, JSON, XML) and decomposes the mapping problem into a richer set of subtasks (e.g., class selection, subject generation, property selection, data reference identification, functions, joins), enabling more fine-grained and format-agnostic evaluation.

Beyond relational databases, the recent SemTab challenge\footnote{\url{https://www.cs.ox.ac.uk/isg/challenges/sem-tab/}} introduces a collection of tabular datasets~\cite{jimenez2020semtab} aimed at automatically matching them to external knowledge graphs such as DBpedia and Wikidata. Several solutions have been proposed, leveraging techniques ranging from heuristic rules and fuzzy matching (e.g., JenTab~\cite{abdelmageed2020jentab}, DAGOBAH\cite{huynh2021dagobah}, and MTAB4D\cite{nguyen2024mtab4d}) to knowledge graph embeddings (e.g, TorchicTab~\cite{dasoulas2023torchictab}). However, SemTab is primarily designed as a benchmark for evaluating end-to-end systems that annotate tables against a pre-existing knowledge graph, rather than aligning them with an ontology. Moreover, most of these systems do not produce explicit, declarative mappings, making it difficult to understand or reproduce how the annotations are actually generated.

In this context, Large Language Models (LLMs) have emerged as a promising solution to automate the transformation of (semi)structured data into KGS, leveraging their advanced natural language understanding capabilities. 
Several recent studies have explored the use of LLMs for generating declarative mapping rules. For instance, Schmidt et al.~\cite{schmidt2025llm} investigate LLM-assisted and context-enhanced YARRRML mapping generation in the manufacturing domain, addressing the challenge of consolidating inventory data in large companies such as Bosch. Their approach supports schema evolution and data integration through semi-automated mapping generation, and they report both qualitative and quantitative evaluations.

In the construction sector, Höltgen et al.~\cite{holtgen2025utilizing} examine how LLMs can be used to convert relational infrastructure data into RDF using a four-step pipeline: SQL query generation, SQL-to-ontology mapping, R2RML mapping generation, and final RDF materialization. Although their results indicate that current LLMs struggle to fully automate this pipeline—due to limitations in handling R2RML syntax and a tendency to hallucinate ontology elements—models like GPT-4o show promising performance, especially under few-shot prompting. In particular, they note the model’s ability to infer implicit relationships in join tables.

Similarly, Hofer et al.~\cite{hofer2024towards} explore the use of LLMs for generating RML files from JSON data in the context of movie knowledge graphs. Their evaluation, using data from the IMDB and a target movie ontology, shows that Claude 3 Opus outperforms GPT-4 in mapping accuracy, while earlier models (e.g., Claude 2.1, GPT-3.5, Gemini-Pro) often produce invalid outputs, highlighting the models’ difficulty in handling precise syntax and semantically constrained tasks.
\begin{table}[t]
    \centering
    \caption{Comparison between RODI, SemTab, and BLINKG in terms of their main focus, evaluation strategy, dependency on existing KGs, supported input formats, granularity and expected mapping outputs.}
    \label{tab:benchmark-comparison}
    \small
    \begin{tabular}{lccc}
        \toprule
        & \textbf{RODI} & \textbf{SemTab} & \textbf{BLINKG} \\
        \midrule
        Main focus
        & RDB--to--ontology
        & Table/column/entity annot.
        & Data--to--ontology \\
        
        Evaluation
        & SPARQL vs.\ SQL
        & Links to KG
        & Explicit Mappings \\

        Granularity
        & Graph-level
        & Annotation-level
        & Task-level \\
        
        Requires existing KG
        & No
        & Yes
        & No \\
        
        Supported formats
        & RDB
        & Tabular data
        & Any \\
        
        Expected Mapping Output
        & R2RML
        & Not explicit
        & Table/RML \\
        
        \bottomrule
    \end{tabular}
\end{table}
Freund et al.~\cite{freund2025mapping} introduce ReMap, a deterministic reverse engineering pipeline to generate RML mappings from non-RDF source data and an expected RDF graph. In their evaluation, mappings generated by LLMs were used as a comparison baseline to assess the performance of ReMap. Additionally, Val-Calvo et al.~\cite{val2025ontogenix} developed OntoGenix, an LLM-powered pipeline for automating ontology development that also generates mappings and RDF data.

While these studies highlight the growing interest in using LLMs for mapping generation, their evaluation strategies remain inconsistent and fragmented. Each proposal defines its own tasks, datasets, and success criteria, making it difficult to compare results or generalize findings. Moreover, most of the evaluations focus primarily on syntactic correctness such as the validity of the generated RML or R2RML files, rather than on the semantic quality or relevance of the mappings. There is a lack of shared tasks, reference datasets, and standardized evaluation parameters. This fragmentation underscores the need for a comprehensive and systematic benchmark like BLINKG, which provides a common ground for evaluating LLM-generated mappings across diverse scenarios, with clear tasks, gold standards, and metrics.

From this discussion, three insights emerge. First, RODI evaluates mappings indirectly, by measuring query-answering quality over the constructed RDF graph, so mapping errors are only observed through their impact on SPARQL results. In contrast, BLINKG focuses on the mapping layer itself: it evaluates the individual mapping decisions (e.g., class, property, subject template, data reference) before materialisation, providing direct evidence of where systems succeed or fail. Second, SemTab assumes an existing knowledge graph and concentrates on entity linking and annotation against that graph, whereas BLINKG targets the prior step of mapping heterogeneous data sources to an ontology, without requiring a pre-populated KG; in this sense, SemTab and BLINKG are complementary, but they address different parts of the pipeline. Third, BLINKG generalises these ideas beyond relational databases to multiple data formats (CSV, JSON, XML) and offers explicit gold standards and metrics for a richer set of mapping-related subtasks, which makes it suitable for systematically evaluating both LLM-based and non-LLM systems under a common, mapping-centric perspective. In Table~\ref{tab:benchmark-comparison}, we present an overview of this comparison between RODI and SemTab with BLINKG.



 
 
 











\section{BLINKG: A Benchmark for LLM-Integrated Knowledge Graph Generation}
\label{sec:blinkg}
To evaluate the performance of automation in mapping generation between ontology and input sources, we designed a benchmark comprising three different scenarios that reflect varying levels of alignment between data sources and ontology schemes: (i) Basic, where column names and ontology terms are lexically similar; (ii) Schema-aligned, where source schemes follow the structure of the ontology but with less obvious lexical overlap; and (iii) Schema-distant, where the source schema is semantically related to the ontology but structurally and lexically distant. These scenarios allow us to systematically test the ability of LLMs to generalize across different mapping difficulties. We also define a set of metrics to evaluate the quality of the generated mappings, considering precision, recall, and F1-score at the level of the task.

\subsection{Knowledge Graph Construction Tasks}

To define the tasks considered in this benchmark, we conducted a thorough review of: i) existing related initiatives, such as SemTab~\cite{jimenez2020semtab}; ii) the capabilities defined by declarative mapping languages, mainly extracted from the complete analysis presented in Iglesias et al.~\cite{iglesias2024ontological} and the mapping features presented in Chaves-Fraga et al.~\cite{chaves2019parameters}. Based on this analysis, we identified the following key tasks:

\begin{itemize}
    \item \textbf{Ontology Class Identification}: This task involves identifying the class of the instance to be generated. For example, if the ontology defines the class \texttt{foaf:Person} and the input data contains a \texttt{person-id}, the output would include a triple such as: \texttt{ex:person/\{id\} rdf:type foaf:Person}. This task is related to the Column Type Annotation (CTA) task proposed in SemTab~\cite{jimenez2020semtab}. However, a key distinction is that input files frequently contain information about multiple types of entities (e.g., denormalized CSV files, XML trees, or JSON documents). Meanwhile, CTA is made at the table level.
    
    \item \textbf{Subject Generation}: This task refers to the generation of the subject in RDF triples. It typically involves two key aspects: i) Following best practices for URI creation~\cite{radulovic2015guidelines}, usually by combining the base URI (e.g., \texttt{http://example.org}) with the class label (e.g., \texttt{Person}); and ii) Identifying data references that ensure instance uniqueness within the dataset, similar to a primary key (PK) in relational databases.
    In this work, we refer to subject generation as \emph{simple} when a single data reference is sufficient to identify an instance, and as \emph{complex} when multiple data references must be combined, analogous to composite primary keys in relational databases.
    While this task is somewhat related to the Cell Entity Annotation (CEA) task in SemTab, it is fundamentally different: CEA focuses on linking a cell value to an existing entity in a KG, whereas the task here is about creating new entities.
    
    \item \textbf{Ontology Property Identification}: This task involves selecting the ontology property that links either a data reference or a related entity class, ensuring the property’s domain matches the identified class. For example, given a column named \texttt{fullname} and an ontology property \texttt{foaf:name} whose domain is \texttt{foaf:Person}, it is very likely that they are semantically equivalent. This task resembles the Column Property Annotation (CPA) task in SemTab; however, our approach leverages ontology restrictions (property domains), whereas CPA relies on already instantiated entities to discover such properties in the target KG.
    
    \item \textbf{Data Reference\footnote{The concept is defined in the RML ontology: \url{http://w3id.org/rml/core/spec\#reference-rml-reference}} Identification}: This task consists of identifying the reference to the input data, such as a CSV column, a JSONPath/XPath expression, or an SQL query, that will be used to construct triples. These references are typically used as the object of a triple when the identified property is a datatype property. Similarly to subject generation, we refer to object generation as \emph{simple} when a single data reference suffices to construct the object value, and as \emph{complex} when multiple data references need to be combined via standard string concatenation to obtain the final literal used in the triple.
    
    \item \textbf{Related Entity Class and Joins}: For object properties defined in the ontology, this task involves specifying how the subject of the related entity (i.e. the range of the property) is generated and under which conditions the relationship is established (i.e., the join condition). For instance, continuing with our earlier example, the triple \texttt{ex:person/1 foaf:works ex:organization/1} would be generated for the object property \texttt{foaf:works} (domain: \texttt{foaf:Person}, range: \texttt{foaf:Organization}) under the condition that \texttt{person.org} equals \texttt{organization.id}.
    
    \item \textbf{Language Annotation.} This task entails assigning the correct language tag to text literals based on their content and context. For each string value in the dataset, such as labels, comments, or descriptions, the system must detect the natural language and annotate it with the appropriate tag (e.g., @en for English, @es for Spanish).
    
    \item \textbf{Datatype Annotation.} This task consists of determining and assigning to each literal the most appropriate RDF datatype IRI, often drawn from XSD but not restricted to it, based on its lexical form and semantic intent. For example, strings matching ISO-8601 date patterns might be annotated as \texttt{xsd:date}, integer numeric values as \texttt{xsd:integer}, boolean-like values as \texttt{xsd:boolean}, or any custom datatype.
    
    \item \textbf{Transformation Functions}: This refers to any transformation that must be applied to data references before triple generation. For instance, dates may need to be transformed from heterogeneous input formats (e.g., \texttt{“26/11/2025”} or \texttt{“11-26-2025”}) into a canonical representation such as \texttt{xsd:date} literals, or boolean-like codes (e.g., \texttt{1}/\texttt{0}, \texttt{“Y”}/\texttt{“N”}) may need to be normalised to explicit \texttt{true}/\texttt{false} values before triple generation. Additionally, one common task is to convert enumerated values (e.g., SQL \texttt{ENUM}s) into standardized taxonomies, such as those represented using SKOS, where each enumerated code is systematically mapped to the URI of the corresponding SKOS concept (e.g., mapping \texttt{‘OPEN’} or \texttt{‘RESTRICTED’} in a procurement system to the appropriate SKOS concepts in a controlled vocabulary). This is particularly relevant in scenarios where KGs leverage standardized datasets (e.g., authority tables from the European Commission that provide standardized taxonomies for public procurement procedures, country names and codes, or currencies\footnote{\url{https://op.europa.eu/en/web/eu-vocabularies/authority-tables}}). These kinds of transformations are aligned with the function mapping capabilities defined in the RML-FNML module\footnote{\url{https://w3id.org/rml/fnml/spec}}, which explicitly supports the specification of such data transformation functions in mapping documents.
\end{itemize}

Although we have split the work into separate tasks for clarity, experts tackle them as a single, intertwined challenge: parsing diverse data formats, mapping fields to the right classes and properties, picking accurate datatypes and language tags, and applying any needed transformations. Pulling this off usually demands deep domain knowledge, meticulous schema analysis, and plenty of careful judgment. Additionally, it would be possible to include more advanced tasks based on the capabilities of newer declarative mapping languages (e.g., RML~\cite{iglesias2023rml}), such as list generation (RML-CC), statements about statements (RML-Star), or advanced input/output descriptions (RML-IO), we decided not to include them in this first version of the benchmark. We consider these to be advanced features, and our current focus is on understanding how proposed solutions perform on the core aspects of KG construction. These more complex features are planned for inclusion in a future version of the benchmark, when more tools are expected to support and process such constructs efficiently.

\subsection{Benchmark Scenarios}
For the BLINKG benchmark, we define three scenarios designed to evaluate the performance of automation proposals, categorizing them by difficulty. Each scenario comprises multiple use cases that adhere to its difficulty level. All use cases within a scenario share a common ontology and controlled vocabularies, ensuring the comparability of the results. Each case is accompanied by one or more datasets in various formats, as well as an expected output that serves as the baseline for evaluation. Table \ref{tab:sce-stats} presents the statistics and data sources of each scenario while Table \ref{tab:sce-overview} provides an overview of the features included in each scenario, which are described in the following sections.

\begin{table}[!t]
\caption{Summary of ontology metrics and input data characteristics.}
\label{tab:sce-stats}
\resizebox{\textwidth}{!}{%
\begin{tabular}{c|c|c|c|c|l}
\cline{1-6}
\textbf{Scenarios} & \textbf{Classes} & \textbf{Data prop.} & \textbf{Object prop.} & \textbf{SKOS}  & \textbf{Input features}\\ \hline
\textbf{1} & 3 & 8 & 3 & 0 & Artificial data based in RML test cases\\
\textbf{2} & 17 & 49 & 10 & 10 & GTFS-Madrid's Metro feed\\
\textbf{3} & 11 & 4 & 10 & 3 & Real life data from CODICE\\ \hline
\end{tabular}
}
\end{table}

\begin{table}[!t]
\caption{Features distributed by each scenario. Scenario 1 is divided in several atomic cases, while Scenarios 2 and 3 represent realistic KG construction scenarios.}
\label{tab:sce-overview}
\resizebox{\textwidth}{!}{%
\begin{tabular}{l|cccccccc|c|c}
\cline{1-11}
\multicolumn{1}{c|}{\multirow{2}{*}{\textbf{Features/Scenarios}}} &
  \multicolumn{8}{c|}{\textbf{Scenario 1}} &
  \multirow{2}{*}{\textbf{\begin{tabular}[c]{@{}c@{}}Scenario 2\\GTFS\end{tabular}}} &
  \multirow{2}{*}{\textbf{\begin{tabular}[c]{@{}c@{}}Scenario 3\\PPDS\end{tabular}}} \\
\multicolumn{1}{c|}{}       & 1A & 1B & 1C & 1D & 1E & 1F & 1G & 1H &   &                      \\ \hline
One data reference          & x  & x  & x  & x  & x  & x  & x  & x  & x & x                    \\
Two or more data references &    & x  & x  & x  & x  & x  & x  & x  & x & x                    \\
Complex object generation   &    &    &    & x  &    & x  & x  &    & x & x                    \\
Simple subject generation   & x  & x  & x  &    & x  &    &    & x  & x & x                    \\
Complex subject generation  &    &    &    & x  &    & x  & x  &    & x & x                    \\
Self join                   &    &    & x  &    &    &    & x  &    & x & x                    \\
Join           &    &    &    &    & x  & x  & x  &    & x & x                    \\
Two or more input sources   &    &    &    &    &    & x  & x  & x  & x & x                    \\
Datatypes generation        &    &    &    & x  &    &    & x  & x  & x & x                    \\
Language annotations        &    &    &    &    &    &    & x  & x  &  &                     \\
Transformation Functions &
  \multicolumn{1}{l}{} &
  \multicolumn{1}{l}{} &
  \multicolumn{1}{l}{} &
  \multicolumn{1}{l}{} &
  \multicolumn{1}{l}{} &
  \multicolumn{1}{l}{} &
  \multicolumn{1}{l}{} &
  \multicolumn{1}{l|}{} &
  x &
  x \\
Distant Schemes             &   &   &   &   &   &   &   &   &  & x \\ \hline
\end{tabular}
}
\end{table}

\paragraph*{Scenario 1: Basic Knowledge Graph Construction}

\begin{figure}[!t]
    \centering
    \includegraphics[width=\textwidth]{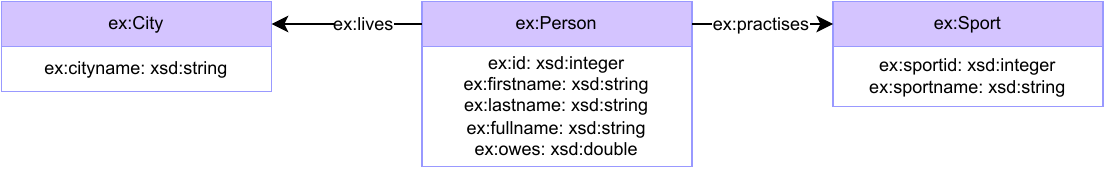}
    \caption{A basic ontology used for the Scenario 1}\label{fig:ontology1}
\end{figure}

This scenario comprises a total of eight atomic cases, and it is inspired by the features defined in RML-core~\cite{RML-core_2023} and RML test cases~\cite{heyvaert2019conformance}. Its objective is to evaluate the creation of KGs for basic use cases with simple input sources. An ontology was created to accommodate all of them, shown in Figure \ref{fig:ontology1}. It is comprised of three classes, related by two object properties, and for each class, there are several data properties.

Input data is provided for each use case in three different formats with the same content: CSV, JSON, and XML. The selection of one format over the other will be an evaluation choice. Regardless of the selected format, the resulting KG will remain the same.

In this scenario, the aim is to test basic and atomic behavior, so all the cases are chosen for this purpose. Case 1A seeks to test the ability to generate a unique mapping, which relates a single column to a data property. The entry refers to the name of an entity whose class is a \texttt{ex:Person}. Since there is insufficient information to conclude whether it is a first name or a last name, the property to be joined with is \texttt{ex:fullname}. As this is the only information available, the subject URI is generated by the same column, acting as a primary key.

Use case 1B increases the complexity of the previous one, performing two mappings on the same entity instead of one, using the same file source in both cases. Each mapping involves a single column and a data property. In addition to the previous use case, the \texttt{ex:id} property would be linked this time, changing the primary key from the column \texttt{Name} to the column \texttt{ID}. As a result, the subject URI would be generated by the \texttt{ID} column.

For use case 1C, two mappings are considered again, coming from the same file, but referring to entities from two different classes, \texttt{ex:Person} and \texttt{ex:Sport}. Thus, each mapping relates to a single data property of each of the entities involved, \texttt{ex:fullname} and \texttt{ex:sportname}. The URIs of the subjects are generated by the corresponding column for each of the entities. By having two entities from different classes, they can be related to each other through an object property, \texttt{ex:practises}, employing a self-join of the triples mapping.

For the 1D use case, four mappings are performed over entities from class \texttt{ex:Person}. Three of them directly relate a column to a data property, such as the first name, last name, or owed amount. The other one links \texttt{ex:fullname} with a literal generated by concatenating the columns for first name and last name. As there is no identifier, the primary key will be established as the concatenation of the first name and last name. The comprehension of the URI concept would be tested, as there are two repeated rows, which refer to the same entity, so the same URI must be considered for both rows. Finally, data types will be checked for consistency with the ontology, as more complex types such as floating-point numbers are included.

For use case 1E, two entities would be involved, \texttt{ex:Person} and \texttt{ex:Sport}, but this time the source files would be different for each entity. Students and sports would be in different files. This does not change the linking of data properties but adds a new difficulty to the establishment of an object property between entities. To achieve that, a conditional join would be necessary that states explicitly the equivalence between the \texttt{Sport} column in the student file and the \texttt{ID} column in the sport file.

Use case 1F explores an alternative way of establishing an object property between entities, by using an additional file that contains the information of the relation. The entities are the same as in the previous use case, but the relation between them is established in a third file. This file contains the ID of the student and the ID of the sport, so two conditional joins would be necessary between the \texttt{ID} column of the student file and the \texttt{Student\_ID} column of the relation file, and between the \texttt{ID} column of the sport file and the \texttt{Sport\_ID} column of the relation file.

For use case 1G, entities from both classes \texttt{ex:Person} and \texttt{ex:City} are considered. The added difficulty in this use case lies in linking the object property \texttt{ex:livesIn} correctly. There are two possible ways of doing it: by a self-join within the same file or by a conditional join between two different files. In this case, unlike the previous ones, the direction to take is not so direct and clear. To generate the mapping, the LLM must be able to avoid being influenced by ambiguity and focus on decision-making. Additionally, the use of language tags for literals is tested, as the city name is a string with language annotations. Although language information is not provided, the context suggests that all city names are in English.

Use case test 1H evaluates the ability to add language annotations to the same data properties of a single entity. Specifically, it involves two instances of the class \texttt{ex:City}, each representing the same city but with names in different languages (English and Spanish). These language differences must be expressed using language tags, not by creating separate entities for each language version. The city identifier serves as the primary key, so the subject URI must remain the same across both language annotations.

In summary, Scenario 1 provides eight atomic cases over a minimal three‐class ontology and plain and simple inputs to validate each core mapping task: from class and subject generation to property identification, joins, datatype, and language annotations. By keeping the setup basic yet varied, we can pinpoint the strengths and weaknesses of automatic pipelines on these foundational building blocks of KG construction.

\paragraph*{Scenario 2: Schema-Aligned Knowledge Graph Construction}
\begin{figure}[!t]
    \centering
    \includegraphics[width=0.9\textwidth]{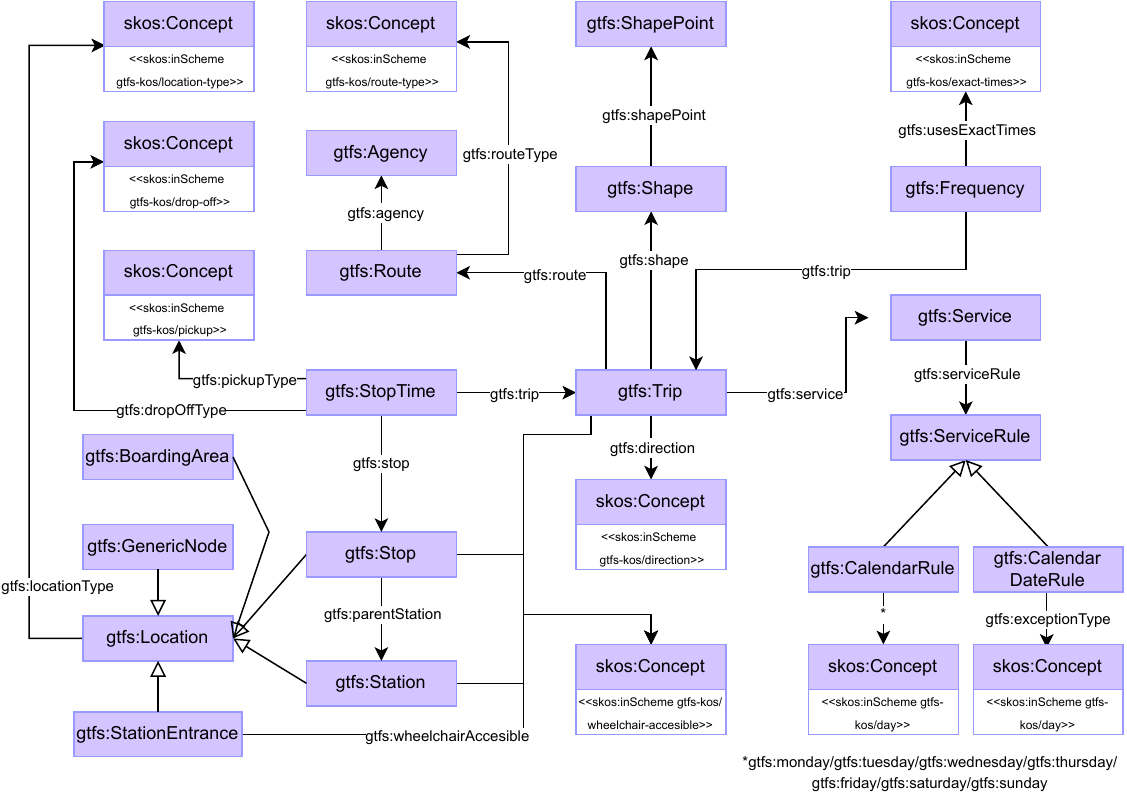}
    \caption{The Linked-GTFS Ontology\footnote{\url{https://github.com/OpenTransport/linked-gtfs}}, used for Scenario 2. We omit datatype properties for the sake of simplicity. White arrows indicate that the source class is an \texttt{rdfs:subClassOf} of the target class, whereas black arrows represent object properties between classes.} 
    \label{fig:ontology2}
\end{figure}

This scenario focuses on building a KG in the transportation domain. We drew inspiration from the GTFS-Madrid-Bench~\cite{chaves2020gtfs} use case, a benchmark designed to evaluate performance throughout the KG construction process. Unlike Scenario 1, which isolates individual tasks, this one presents a more realistic and complex setting where those tasks frequently overlap and interact. However, the main feature of this scenario is the very high alignment between the input data and the ontology. The ontology itself was built bottom-up directly from the official GTFS specification\footnote{\url{https://gtfs.org}}. In other words, the ontology can be seen as a near-mirror of the GTFS specification, yet it diverges in important ways as OWL and RDF supports richer modeling constructs (e.g. complex class hierarchies, property restrictions, controlled vocabularies) that GTFS’s flat, tabular-based format cannot express.

The goal of this scenario is to assess whether automated solutions operating in a context where input schemes and the ontology closely align can still understand ontology peculiarities, such as richer modeling constructs, for correctly transforming the input data into a KG. From a technical perspective, it has the objective of testing, in addition to everything included in Scenario 1 (see Table \ref{tab:sce-overview}), the handling of transformation functions and the management of a larger amount of data, as in a more realistic use case. The ontology considered is represented in Figure \ref{fig:ontology2}, and has seventeen classes that are described in Table \ref{tab:scenario2}. The input data are 10 files extracted from the GTFS-Madrid-Bench~\cite{chaves2020gtfs} database\footnote{Note that the benchmark provides support for different data formats such as RDB, CSV, XML, or JSON}. Two examples of input sources (stops and routes) are shown in Listing~\ref{lst:stops} and Listing~\ref{lst:routes}. The information contained in them covers the entire set of classes and properties considered in this scenario.

\begin{table}[!t]
    \caption{Class descriptions for ontology of Scenario 2 } \label{tab:scenario2}
    \begin{tabular}{@{} l p{0.75\textwidth} @{}}
    \hline
    \textbf{Class} & \textbf{Description} \\
    \hline
    Agency         & Transit companies with service. \\
    BoardingArea   & Location where passengers can board and/or alight vehicles. \\
    CalendarRule   & Service dates specified using a weekly schedule with start and end dates. \\
    CalendarDateRule & Exception dates for the services. \\
    Frequency      & Trip gap for frequency service or condensed schedule. \\
    GenericNode    & A location within a station, not matching any other location type. \\
    Location       & Places where vehicles pick up or drop off riders. \\
    Route          & Group of trips that are displayed to riders as a single service. \\
    Service        & Set of ServiceRules. \\
    ServiceRule    & Rule that associates dates with services (CalendarRule/CalendarDateRule). \\
    Shape          & Rules for mapping vehicle travel paths. \\
    ShapePoint     & One point in a Shape. \\
    Station        & Large transit location that may contain multiple Stops. \\
    StationEntrance& Location where passengers can enter or exit a station from the street.\\
    Stop           & Physical location where a vehicle stops or leaves. \\
    StopTime       & Times that a vehicle arrives at and departs from stops for each trip. \\
    Trip           & Sequence of two or more stops that occur during a specific time period. \\
    \hline
  \end{tabular}
\end{table}

The main difference from Scenario 1 is the introduction of more complexity at the ontology level, such as class hierarchies and constraints on classes and properties. 
The ontology introduces two distinct hierarchies: one for describing locations (stops, stations, station entrances, boarding areas, and generic nodes) and another for modeling service calendars. The calendar hierarchy maps directly to GTFS, since the specification defines two separate input files (\texttt{calendar.txt} and \texttt{calendar\_dates.txt}), each corresponding to one of these classes. In contrast, the location hierarchy depends on the value of the \texttt{location\_type} column in \texttt{stops.txt} (see Listing~\ref{lst:stops}), adding complexity to the mapping rules for these subclasses. Notably, the same \texttt{location\_type} value also determines inter-location relationships via the \texttt{parent\_station} property, which entails a self-join on \texttt{stops.txt} to generate different subclasses.  

\begin{lstlisting}[%
  caption={Excerpt of the input source stops in CSV},%
    float,
  floatplacement=thbp,
  label={lst:stops},%
  basicstyle=\ttfamily\footnotesize,
  linewidth=\textwidth%
]
stop_id,stop_code,name,stop_desc,lat,lon,location_type,parent_station,wheelchair_boarding
acc411,1,Plaza de Castilla,P. de la Castellana 189,40.46682,-3.68918,0,est9021,0
acc41040,1,Ascensor,Pz. de Castilla 9,40.46555,-3.68877,2,est9021,1
acc41043,1,Intercambiador Superficie,P. de la Castellana 191 B,40.46728,-3.68915,2,est9021,0
acc41044,1,Ascensor,P. de la Castellana 189,40.46702,-3.68918,2,est9021,0
acc412,1,Castellana impares,P. de la Castellana 189,40.46722,-3.68952,2,est_90_21,0
acc413,1,Bravo Murillo,C. de Bravo Murillo 377,40.46629,-3.69036,1,est_90_21,1
\end{lstlisting}

\begin{lstlisting}[%
  caption={Excerpt of the input source routes in CSV},%
  label={lst:routes},%
    float,
  floatplacement=thbp,
  basicstyle=\ttfamily\footnotesize,
  linewidth=\textwidth
]
route_id,agency_id,route_short_name,route_long_name,route_type,route_color,route_text_color
41,CRTM,1,Pinar de Chamartín-Valdecarros,1,2DBEF0,FFFFFF
410,CRTM,10,Hospital del Norte-Puerta del Sur,1,005AA9,FFFFFF
411,CRTM,11,Plaza Elíptica-La Fortuna,1,009B3A,FFFFFF
412,CRTM,12,MetroSur,1,A49800,FFFFFF
\end{lstlisting}

In terms of constraints, many properties such as \texttt{gtfs:id}, apply to multiple classes, forcing its domain to be defined as the union of those classes and increasing the complexity of linking classes to input sources. We also propose the use of value restrictions on datatype properties. For example, the \texttt{gtfs:longName} of class \texttt{gtfs:Route} must start with an uppercase letter and consist of an alphanumeric string. 
All such restrictions may be encoded as transformation functions that ensure incoming data conforms to the ontology’s rules before KG generation.

Finally, the scenario incorporates controlled vocabularies. When input data contains enumerated types (data types defined by a fixed set of named values), these are usually transformed into independent taxonomies outside the ontology, often leveraging the W3C SKOS standard vocabulary\footnote{\url{https://www.w3.org/TR/skos-reference/}}. By extracting the enumerated fields from GTFS, we generated eleven controlled vocabularies: Wheelchair Accessibility, Location Type, Route Type, Pickup, Drop‐off, Direction ID, Bikes Allowed, Timepoint, Service Availability, Exception Type, and Exact Times. The essential task in this phase is to interpret each input value and map it to the corresponding concept within its scheme. For example, GTFS defines three numeric codes for bicycle access (0—no information; 1—allowed; 2—not allowed), and our controlled vocabulary represents these as URIs: wheel-chair-kos:no‐information, wheel-chair-kos:no‐information:accessible, wheel-chair-kos:no‐information:inaccessible\footnote{\texttt{wheel-chair-kos:} is the prefix of http://transport.linkeddata.es/kos/wheelchair-accesible/}.
The main challenge is understanding the meaning of these codes (often provided without context) and assigning them to the correct taxonomy entries. In the KG construction pipeline, this is can be handled by transformation functions that convert raw input data into valid controlled vocabulary values.

This scenario brings everything together in a more realistic setting: a GTFS-based ontology that mirrors the standard but adds OWL-style class hierarchies, property constraints, and controlled vocabularies. Beyond the tasks from Scenario 1, we now also put transformation functions and a more complex environment to the test. The goal is to see if automated pipelines can tackle these added challenges and still build a robust KG.

\paragraph*{Scenario 3: Schema-Distant Knowledge Graph Construction}

\begin{figure}[!t]
    \centering
    \includegraphics[width=0.9\textwidth]{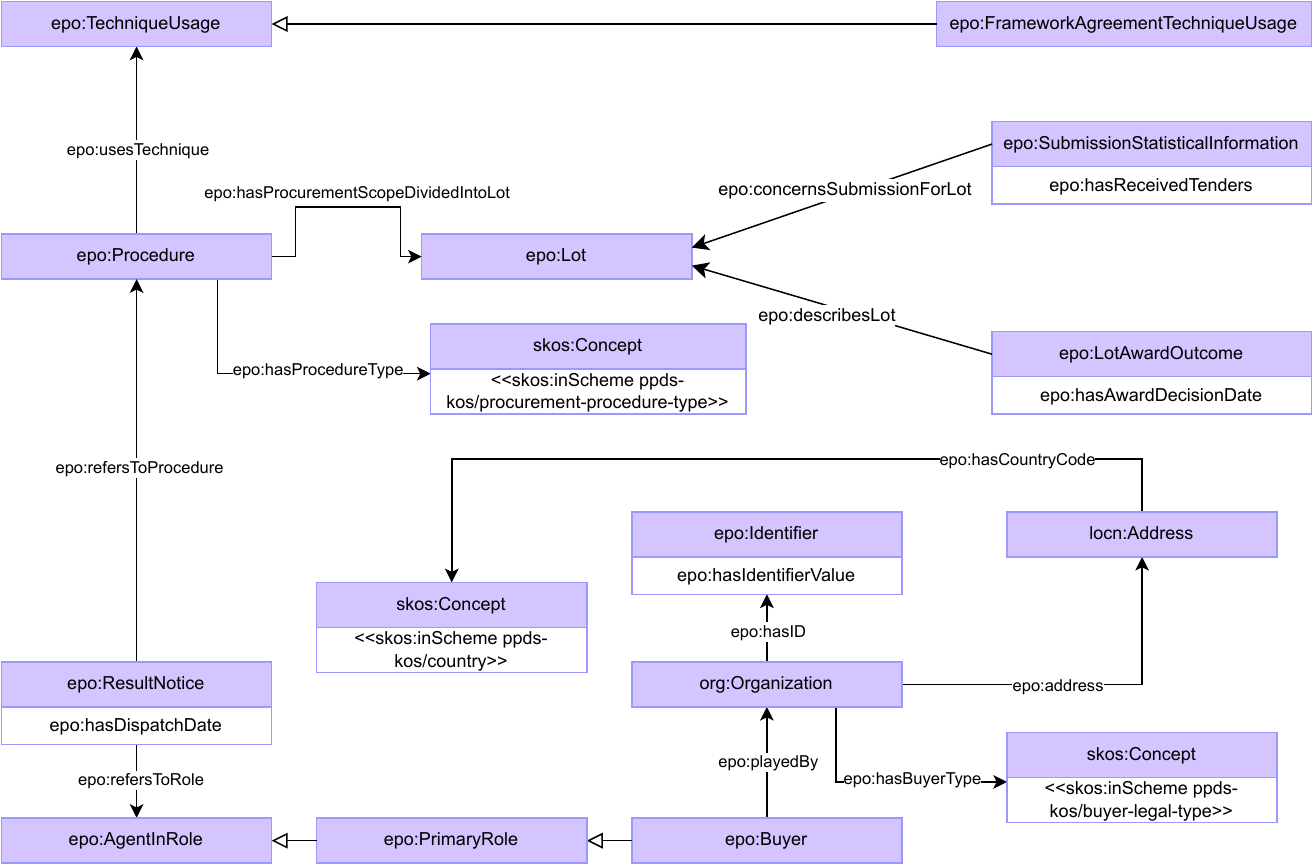}
    \caption{The subset of the e-Procurement Ontology used for the Scenario 3. White arrows indicate that the source class is an \texttt{rdfs:subClassOf} of the target class, whereas black arrows represent object properties between classes.}\label{fig:ontology3}
\end{figure}

This scenario is the most complex scenario, focusing on building a KG for public procurement. It follows Scenario 2’s methodology but pushes it further by operating in a realistic setting where tasks frequently interact and overlap. The key difference is that, unlike Scenario 2, the input data models and the ontology are not closely aligned. Although they cover the same domain and share similar concepts, their schemes and lexical descriptions diverge significantly, meaning that deep domain expertise is required to create the relationships between the input sources and the provided ontology.

The ontology for Scenario 3 is drawn from the eProcurement Ontology (ePO)\footnote{\url{https://github.com/OP-TED/ePO}}, the official, pan-European data model for public procurement maintained by the European Publications Office. Given the domain’s complexity, ePO is organized into 13 ontology modules, each targeting a specific aspect of the procurement lifecycle, from contract notices and award procedures to financial executions and legal frameworks. For our purposes, we extracted a focused subset of classes from the epo-core module, which alone defines nearly 150 classes and is under active, continuous development.This ontology supports multiple initiatives, such as the EU Public Procurement Data Space (PPDS)\footnote{\url{https://single-market-economy.ec.europa.eu/single-market/public-procurement/digital-procurement/public-procurement-data-space-ppds_en}}, which aims to harmonize heterogeneous procurement datasets from all EU Member States into a common semantic layer based on ePO. A key objective of PPDS is to support the standardized computation and comparison of public procurement transparency indicators, as published in the Single Market Scoreboard\footnote{\url{https://single-market-scoreboard.ec.europa.eu/business-framework-conditions/public-procurement}}. Moreover, each Member State typically operates its own national procurement platform, with distinct data formats, schemas, and publication workflows. As a result, manually integrating all these sources with respect to the ePO ontology is extremely costly, error-prone, and must be tailored to the specific characteristics of each country.

\begin{table}[!t]
    \caption{Class descriptions for ontology of Scenario 3 } \label{tab:scenario3}
    \resizebox{\textwidth}{!}{%
    \begin{tabular}{@{} l p{0.75\textwidth} @{}}
    \hline
    \textbf{Class} & \textbf{Description} \\
    \hline
    AgentInRole   & Ties an agent to a part they play in a given situational context. \\
    Buyer & Role of an agent that awards a contract and/or purchases items.\\
    FrameworkAgreementTechniqueUsage & Technique that establishes the terms governing contracts to be awarded.\\
    Identifier & String to distinguish uniquely one instance of an object.\\
    Lot & Division of the 
    services 
    to be procured, allowing the award of contracts.\\
    LotAwardOutcome & Result concerning the Lot attributed by the awarder.\\
    PrimaryRole & A primary role within the procurement process that ties an agent to a part.\\
    Procedure & Set of administrative activities conducted to conclude one or more contracts.\\
    ResultNotice & Announcement of the award or non-award of a contract by a buyer.\\
    SubmissionStatisticalInformation & Statistical information about submissions on a given Lot.\\
    TechniqueUsage & Methods used for conducting procurement procedure.\\
    \hline
  \end{tabular}
  }
\end{table}

The ontology used in this scenario is shown in Figure \ref{fig:ontology3}. It comprises 13 classes that capture details of public procurement processes, their outcomes, award procedures, and the organizations involved as contractors. Their definitions are also provided in Table \ref{tab:scenario3}. Furthermore, each procurement process can be divided into multiple lots, each of which may follow a distinct awarding procedure. From a technical standpoint, this ontology features deep, nested class hierarchies such as \texttt{epo:Buyer} → \texttt{epo:PrimaryRole} → \texttt{epo:AgentRole} and, despite a reduction in datatype properties compared to Scenario~2, it defines a large set of object properties to capture the rich relationships inherent in the public procurement domain. Nevertheless, we observe repetitive restriction patterns w.r.t. Scenario~2, such as properties reused across multiple classes and explicit cardinality constraints imposed on certain classes. Regarding controlled vocabularies, ePO reuses those provided by the Publications Office (the so-called authority tables\footnote{\url{https://op.europa.eu/en/web/eu-vocabularies/authority-tables}}), so input data must be mapped and linked to these external taxonomies. As shown in Figure~\ref{fig:ontology3}, this scenario leverages three controlled vocabularies: Procedure Types\footnote{\url{http://publications.europa.eu/resource/authority/procurement-procedure-type}}, Country Codes\footnote{\url{http://publications.europa.eu/resource/authority/country}}, and Buyer Types\footnote{\url{http://publications.europa.eu/resource/authority/buyer-legal-type}}.

\begin{lstlisting}[
  language=XML,
  float,
  floatplacement=tp,
  basicstyle=\ttfamily\footnotesize,
  breaklines=true,
  caption={Excerpt from CODICE, the Public Procurement Spanish Platform},
  label={lst:ppds-entry},
  linewidth=\textwidth
]
<entry>
  <id>https://contrataciondelestado.es/.../6854467</id>
  <title>Suministro de piezas de recambio [...] EMAYA.</title>
  <updated>2021-01-07T10:47:01.782+01:00</updated>
  <summary type="text">
    Id licitacion: 942P LOTE 2; Organo de Contratacion: Gerencia de EMAYA
  </summary>
  <cac-place-ext:ContractFolderStatus>
    <cac:Party>
      <cbc:WebsiteURI>http://www.emaya.es</cbc:WebsiteURI>
      <cac:PartyIdentification>
        <cbc:ID schemeName="NIF">A07000029</cbc:ID>
      </cac:PartyIdentification>
      <cac:PartyName>
        <cbc:Name>Gerencia de EMAYA, Empresa Municipal [...]</cbc:Name>
      </cac:PartyName>
      <cac:PostalAddress>
        <cbc:CityName>Palma</cbc:CityName>
        <cbc:PostalZone>07010</cbc:PostalZone>
        <cac:AddressLine>
          <cbc:Line>
            Camino de los Reyes 400, Edificio Central de Son Pacs
          </cbc:Line>
        </cac:AddressLine>
      </cac:PostalAddress>
    </cac:Party>
  </cac-place-ext:ContractFolderStatus>
 <!--...and more nested tags...-->
</entry>
\end{lstlisting}

The input data for this scenario comes from the Spanish public procurement platform CODICE\footnote{\url{https://contrataciondelestado.es/wps/portal/codice}}. CODICE supports the entire procurement lifecycle and provides a suite of standard XSD schemas to ensure interoperability across all public administrations. Listing~\ref{lst:ppds-entry} shows a brief XML excerpt of these input files, conforming to the official CODICE XSD definitions. The main challenge in transforming this data into a KG based on the ePO ontology lies in the complex, deeply nested XML tree: entities corresponding to each ontology class and their properties are scattered throughout the document. Consequently, the pipeline must parse each branch independently, leveraging element names and attribute values to instantiate ontology classes correctly. Since CODICE and ePO were developed independently (they cover the same domain but lack the high alignment of Scenario 2), a thorough understanding of both models is crucial to establish accurate mappings. From a technical point of view, this scenario incorporates all the challenges already presented in Scenario~2 and adds an extra layer of complexity when defining join conditions for object‐property triples. In some cases, these joins must span different levels of the XML tree, linking elements on separate branches, which requires robust mapping logic to correlate related entities accurately.

In summary, Scenario 3 tests the full spectrum of challenges: from multi-level XML parsing and independent schema-ontology alignment to nested joins and controlled-vocabulary linking—within a real-world public procurement setting. By combining the rich, modular ePO ontology with CODICE’s complex XML structures and national variations, this scenario pushes automated KG pipelines to demonstrate both domain expertise and flexible mapping strategies under realistic conditions.

\subsection{Metrics and Expected Output}

Reliable, quantitative metrics are essential for evaluating how well automated systems handle each of the identified tasks involved in KG construction. Defining a fair metric is complex: semantic equivalence can be expressed in many ways, and simple string matching misses valid paraphrases or alternate URIs. It is also known that LLMs struggle with syntax problems when generating RML or SPARQL-Anything rules~\cite{hofer2024towards,maushagen2024populating}, so BLINKG does not focus on evaluating whether systems can produce syntactically valid mapping languages. Instead, for each scenario we ask the model to produce a structured, tabular output that explicitly lists, for example, the selected classes, properties, and data references for each column or field. A detailed description of the expected output is shown in Table \ref{tab:mapping} and Table \ref{tab:expected} shows illustrative example of this expected output format extracted from the Scenario 2. Thus, for evaluation, we frame each identified task (Ontology Class Identification, Subject URI Generation, Ontology Property Identification, etc.) as a binary outcome: the model’s output is either correct or incorrect. By breaking the pipeline into these subtasks, BLINKG can measure Precision, Recall, and F-score on each task independently.

\begin{table}[!t]
  \centering
  \caption{Expected output table from the execution of each experiment.}
  \label{tab:mapping}
  \begin{tabular}{@{} l p{0.75\textwidth} @{}}
    \hline
    \textbf{Column} & \textbf{Description} \\
    \hline
    Data Reference        & Each data reference from the input data source files. \\
    Ontology Property     & Name of the ontology property related with the column. \\
    Entity Class          & Class of the subject considered. \\
    Related Entity Class  & Class of the corresponding object. \\
    Subject Generation    & Template to generate the subject URI. \\
    Join & Equivalence condition in a join. \\
    Datatype              & Object's datatype. \\
    Language Annotations  & Object's language tag. \\
    Function Name         & Name of the function needed. \\
    Function Output       & Transformation applied to data reference using the function. \\
    \hline
  \end{tabular}
\end{table}

\begin{table}[!t]
\centering
\caption{Example of the expected output from Shape and ShapePoint classes from the LinkedGTFS ontology and the shapes.txt file from GTFS spec. All ontology references starting from \texttt{:} mean that are using the base ontology IRI, while all subjects should start with \texttt{ex:shape/}. Functions are not shown due to the lack of space}.
\label{tab:expected}
\resizebox{\textwidth}{!}{%
\begin{tabular}{ccccccc}
\hline
\textbf{Data Ref.} &
  \textbf{Property} &
  \textbf{Class} &
  \textbf{Rel. Class} &
  \textbf{Subject (ex:shape/)} &
  \textbf{Join Con.} &
  \textbf{Datatype} \\ \hline
shape\_id &
:id &
:Shape &
   &
\{shape\_id\} &
   &
xsd:string  \\ \hline
shape\_pt\_lat &
:latitude &
:ShapePoint &
   &
  \begin{tabular}[c]{@{}c@{}}\{shape\_id\}-\\ \{shape\_pt\_seq\}\end{tabular} &
   &
geo:lat  \\ \hline
shape\_pt\_lon &
:longitude &
:ShapePoint &
   &
  \begin{tabular}[c]{@{}c@{}}\{shape\_id\}-\\ \{shape\_pt\_seq\}\end{tabular} &
   &
geo:lon  \\ \hline
shape\_pt\_seq &
:pointSequence &
:ShapePoint &
   &
  \begin{tabular}[c]{@{}c@{}}\{shape\_id\}-\\ \{shape\_pt\_seq\}\end{tabular} &
   &
xsd:integer  \\ \hline
shape\_dist &
:distanceTraveled &
:Shape &
   &
\{shape\_id\} &
   &
xsd:float  \\ \hline
shape\_id &
:shapePoint &
:Shape &
:ShapePoint &
\{shape\_id\} &
  \begin{tabular}[c]{@{}c@{}}shape\_id = \\ shape\_id\end{tabular} &
   \\ \hline
\end{tabular}%
}
\end{table}

We quantify performance on each KG‐construction task using precision, recall and F‐score. Precision measures the fraction of an approach’s correct outputs, recall is the fraction of expert‐validated items the model recovers, and F-score balances the two as their harmonic mean. Applying these metrics to each binary task outcome yields a clear, repeatable assessment of model strengths and weaknesses.


Since relying on a single execution of a prompt may lead to unstable or biased results, we propose to run the same prompt multiple times under identical conditions. We then compute a macro-average of all metrics over these runs, where \texttt{n} denotes the number of evaluations of the same prompt:

$$
\textit{Macro-Precision}= \frac{1}{n} \sum_{i=1}^{n} Precision_i,\quad 
\textit{Macro-Recall}= \frac{1}{n} \sum_{i=1}^{n} Recall_i $$
$$
\textit{Macro-F-score}= \frac{1}{n} \sum_{i=1}^{n} \textit{F-score}_i $$

Although these metrics provide good support for quantifying the quality of the results, relying on a manual evaluation is not scalable, and strict string-equality checks fall short: an automatic approach might express the same output with a paraphrase, synonym, or alternate URI that could not match the reference ground truth. Similar to what is done in other domains such as NLP, we enhance precision, recall, and F-score by inserting a preliminary similarity check between each output and the ground truth. We compute three complementary measures:
\begin{itemize}
    \item Levenshtein distance for a straightforward string‐level similarity score.
    \item Cosine similarity over a language model embeddings (such as SBERT) of the raw output, capturing semantic closeness beyond exact text matches.
    \item Cosine similarity on ontology‐driven verbalizations, where we replace the raw output with the canonical label or description of the target class/property (retrieved via lookup) and computing the embeddings on that text.
\end{itemize}

Because the identified tasks range from simple class identification to complex joins, we compute all three similarity scores (Levenshtein, raw‐embedding cosine and verbalization‐embedding cosine) for each candidate mapping and take the highest value. We then compare that top score against a predefined threshold: if it exceeds the threshold, the annotation is marked correct; otherwise, it’s deemed incorrect. In detail, the proposed metric would be:

$$
\text{correct annotation} =
\begin{cases}
1, & \max\bigl\{\,s_{\mathrm{lev}},\;s_{\mathrm{raw}},\;s_{\mathrm{verbal}}\,\bigr\} \ge \tau,\\
0, & \max\bigl\{\,s_{\mathrm{lev}},\;s_{\mathrm{raw}},\;s_{\mathrm{verbal}}\,\bigr\} < \tau
\end{cases}
$$

Where $s_{\mathrm{lev}}$ is the normalized Levenshtein similarity, $s_{\mathrm{raw}}$ is the cosine similarity with raw-input embeddings, $s_{\mathrm{verbal}}$ is the cosine similarity of verbalized‐ontology embeddings, and $\tau$ is the chosen threshold.

\subsection{Sustainability and Community}
BLINKG is designed as a living, community-driven benchmark rather than a one-off snapshot. All resources (ontologies, datasets, gold standards, evaluation library, and examples) are released in an open repository\footnote{\url{https://github.com/citiususc/blinkg}}, so that scenarios and baselines can evolve over time while preserving reproducibility of past results. The benchmark is proposed as method-agnostic: although in the paper we are focused on testing LLM-based approaches, the same scenarios and evaluation pipeline can be used to assess rule-based systems, hybrid pipelines, or human-in-the-loop workflows. We provide detailed documentation, example configurations, to lower the barrier for its adoption.

To foster community engagement, we explicitly invite contributions of new scenarios, baselines, and results through GitHub issues, pull requests, and discussions. We provide dedicated issue templates and guidelines for submitting additional resources, so that contributors can extend BLINKG in a structured and consistent way (e.g., by covering new domains, data formats, levels of schema distance, or mapping subtasks). We also plan to align future iterations of BLINKG with the Knowledge Graph Construction Challenge\footnote{\url{http://w3id.org/kg-construct/workshop/\#challenge}}, where the benchmark can be used in new tracks (e.g., automatic KG Construction) and enriched with more demanding configurations, including scenarios based on synthetic ontologies and input data. This tight integration between an open benchmark, a public repository, and a recurring community event is intended to ensure the long-term sustainability, relevance, and extensibility of BLINKG.

\section{Evaluation}
\label{sec:evaluation}

In this section, we outline the BLINKG evaluation in three steps. First, we introduce the selected LLMs. Next, we detail our evaluation procedure and methodology. Finally, we present and discuss the results. 

\subsection{Selected LLMs}
Several promising LLMs have been selected from the state-of-the-art, combining both open-access and fee-based options, as well as reasoning models. This selection was motivated by the use of these models in previous works~\cite{freund2025mapping,hofer2024towards,holtgen2025utilizing}, which show medium-to-high performance, aiming for a balanced representation of both proprietary and open-source LLMs. In all our experiments, we set the LLM temperature to 0.3, as preliminary runs with different values indicated that this configuration provided the best trade-off between stability and diversity in the generated mappings.

\textbf{DeepSeek-R1.} DeepSeek-R1~\cite{deepseekai2025}, developed by DeepSeek AI in 2025, is a reasoning foundational model trained exclusively through large-scale reinforcement learning, eliminating the need for supervised fine-tuning. This approach allows it to develop strong reasoning and problem-solving skills by learning directly from interactions, optimizing for adaptability and long-term outcomes. Its architecture excels at complex tasks such as logical inference and decision-making.

\textbf{Gemini 2.5 Pro.} Gemini 2.5 Pro~\cite{geminiteam2024}, developed by Google in 2025, is a multimodal foundational model that excels in language understanding, reasoning, and complex task resolution. It integrates text, images, audio, and video data for deeper contextual insights, enabling accurate inferences and efficient problem-solving. Optimized for scalability and adaptability, it leverages self-supervised learning and domain-specific fine-tuning. 

\textbf{GPT-4 Omni.} GPT-4o~\cite{openai2024} is an autoregressive foundational model developed by OpenAI in 2024. It is a multimodal model, capable of processing and generating text, images, and audio within a unified architecture. This model excels in complex problem-solving and contextual understanding. Additionally, it achieves high proficiency in code generation, producing accurate and efficient code across multiple programming languages. 

\textbf{OpenAI o3.} OpenAI o3~\cite{openai2025o3} is a reasoning foundational model developed by OpenAI in 2025, as an evolution of OpenAI o1~\cite{openai2024o1}, which was trained with large-scale reinforcement learning to reason using a chain of thought. As a result of this targeted training, GPT-o3 demonstrates advanced reasoning abilities across a wide range of cognitive tasks, including complex problem-solving and logical inference.

\textbf{LLaMa 3.3 70B Instruct.} Llama 3.3~\cite{grattafiori2024} is an instruction-tuned generative model in 70B released by Meta in 2024. It is an autoregressive language model that uses an optimized transformer architecture. The tuned versions use supervised fine-tuning and reinforcement learning with human feedback to align with human preferences for helpfulness and safety. LLaMa 3.3 70B Instruct is an instruct fine-tuned version of LLaMa 3.3 70B.

\textbf{Mixtral 8x22B Instruct.} Mixtral 8x22B~\cite{mistral2024mixtral} is a foundational model developed by Mistral AI. It is a sparse Mixture-of-Experts (SMoE) model that uses only 39B active parameters out of 141B, offering unparalleled cost efficiency for its size. It delivers good performance in complex natural language processing tasks, such as text comprehension, content generation, and advanced reasoning, rivaling larger models while reducing computational costs. Its ability to handle extensive contexts and adapt to various domains makes it a versatile tool. Mixtral 8x22B Instruct is an instruct fine-tuned version of Mixtral 8x22B.

\begin{table}[!t]
\centering
\caption{Statistics for the ontology and input data used in each scenario, together with the expected number of task instances in the gold standard.}
\label{tab:experiments}
\resizebox{\textwidth}{!}{%
\begin{tabular}{cc|cccccccc|c|c}

\multicolumn{2}{c|}{\multirow{2}{*}{}} &
  \multicolumn{8}{c|}{\textbf{Scenario1}} &
\multirow{2}{*}{\textbf{\begin{tabular}[c]{@{}c@{}}Scenario 2\\ GTFS\end{tabular}}} &
  \multirow{2}{*}{\textbf{\begin{tabular}[c]{@{}c@{}}Scenario 3\\ PPDS\end{tabular}}} \\ \cline{3-10}
\multicolumn{2}{c|}{} &
  \multicolumn{1}{c|}{\textbf{1A}} &
  \multicolumn{1}{c|}{\textbf{1B}} &
  \multicolumn{1}{c|}{\textbf{1C}} &
  \multicolumn{1}{c|}{\textbf{1D}} &
  \multicolumn{1}{c|}{\textbf{1E}} &
  \multicolumn{1}{c|}{\textbf{1F}} &
  \multicolumn{1}{c|}{\textbf{1G}} &
  \textbf{1H} &
   &
   \\ \hline
\multicolumn{1}{c|}{\multirow{10}{*}{\textbf{Task}}} &
  \textbf{Class} &
  \multicolumn{1}{c|}{1} &
  \multicolumn{1}{c|}{1} &
  \multicolumn{1}{c|}{2} &
  \multicolumn{1}{c|}{1} &
  \multicolumn{1}{c|}{2} &
  \multicolumn{1}{c|}{2} &
  \multicolumn{1}{c|}{2} &
  1 &
  12 &
  8 \\ \cline{2-12} 
\multicolumn{1}{c|}{} &
  \textbf{SubjectGen} &
  \multicolumn{1}{c|}{1} &
  \multicolumn{1}{c|}{1} &
  \multicolumn{1}{c|}{2} &
  \multicolumn{1}{c|}{1} &
  \multicolumn{1}{c|}{2} &
  \multicolumn{1}{c|}{2} &
  \multicolumn{1}{c|}{2} &
  1 &
  10 &
  13 \\ \cline{2-12} 
\multicolumn{1}{c|}{} &
  \textbf{Prop.} &
  \multicolumn{1}{c|}{1} &
  \multicolumn{1}{c|}{2} &
  \multicolumn{1}{c|}{3} &
  \multicolumn{1}{c|}{4} &
  \multicolumn{1}{c|}{5} &
  \multicolumn{1}{c|}{7} &
  \multicolumn{1}{c|}{6} &
  1 &
  42 &
  16 \\ \cline{2-12} 
\multicolumn{1}{c|}{} &
  \textbf{DataRef} &
  \multicolumn{1}{c|}{1} &
  \multicolumn{1}{c|}{2} &
  \multicolumn{1}{c|}{2} &
  \multicolumn{1}{c|}{4} &
  \multicolumn{1}{c|}{5} &
  \multicolumn{1}{c|}{7} &
  \multicolumn{1}{c|}{6} &
  3 &
  44 &
  7 \\ \cline{2-12} 
\multicolumn{1}{c|}{} &
  \textbf{Rel.Entity} &
  \multicolumn{1}{c|}{} &
  \multicolumn{1}{c|}{} &
  \multicolumn{1}{c|}{1} &
  \multicolumn{1}{c|}{} &
  \multicolumn{1}{c|}{1} &
  \multicolumn{1}{c|}{1} &
  \multicolumn{1}{c|}{1} &
   &
  10 &
  16 \\ \cline{2-12} 
\multicolumn{1}{c|}{} &
  \textbf{Joins} &
  \multicolumn{1}{c|}{} &
  \multicolumn{1}{c|}{} &
  \multicolumn{1}{c|}{1} &
  \multicolumn{1}{c|}{} &
  \multicolumn{1}{c|}{1} &
  \multicolumn{1}{c|}{1} &
  \multicolumn{1}{c|}{1} &
   &
  9 &
  16 \\ \cline{2-12} 
\multicolumn{1}{c|}{} &
  \textbf{Lang.Tag} &
  \multicolumn{1}{c|}{} &
  \multicolumn{1}{c|}{} &
  \multicolumn{1}{c|}{} &
  \multicolumn{1}{c|}{} &
  \multicolumn{1}{c|}{} &
  \multicolumn{1}{c|}{} &
  \multicolumn{1}{c|}{1} &
  2 &
   &
   \\ \cline{2-12} 
\multicolumn{1}{c|}{} &
  \textbf{Datatype} &
  \multicolumn{1}{c|}{} &
  \multicolumn{1}{c|}{} &
  \multicolumn{1}{c|}{} &
  \multicolumn{1}{c|}{2} &
  \multicolumn{1}{c|}{} &
  \multicolumn{1}{c|}{} &
  \multicolumn{1}{c|}{3} &
   &
  14 &
  3 \\ \cline{2-12} 
\multicolumn{1}{c|}{} &
  \textbf{Functions} &
  \multicolumn{1}{c|}{} &
  \multicolumn{1}{c|}{} &
  \multicolumn{1}{c|}{} &
  \multicolumn{1}{c|}{} &
  \multicolumn{1}{c|}{} &
  \multicolumn{1}{c|}{} &
  \multicolumn{1}{c|}{} &
   &
  10 &
  3 \\ \cline{2-12} 
\multicolumn{1}{c|}{} &
  \textbf{Total} &
  \multicolumn{1}{c|}{1} &
  \multicolumn{1}{c|}{2} &
  \multicolumn{1}{c|}{3} &
  \multicolumn{1}{c|}{4} &
  \multicolumn{1}{c|}{5} &
  \multicolumn{1}{c|}{7} &
  \multicolumn{1}{c|}{6} &
  3 &
  72 &
  26 \\ \hline
\multicolumn{2}{c|}{\textbf{OWL Ontology}} &
  \multicolumn{1}{c|}{AdHoc} &
  \multicolumn{1}{c|}{AdHoc} &
  \multicolumn{1}{c|}{AdHoc} &
  \multicolumn{1}{c|}{AdHoc} &
  \multicolumn{1}{c|}{AdHoc} &
  \multicolumn{1}{c|}{AdHoc} &
  \multicolumn{1}{c|}{AdHoc} &
  AdHoc &
  LinkedGTFS &
  Subset of ePO \\ \hline
\multicolumn{2}{c|}{\textbf{SKOS}} &
  \multicolumn{1}{c|}{} &
  \multicolumn{1}{c|}{} &
  \multicolumn{1}{c|}{} &
  \multicolumn{1}{c|}{} &
  \multicolumn{1}{c|}{} &
  \multicolumn{1}{c|}{} &
  \multicolumn{1}{c|}{} &
   &
  10 &
  3 \\ \hline
\multicolumn{2}{c|}{\textbf{Input Data}} &
  \multicolumn{1}{c|}{1 CSV} &
  \multicolumn{1}{c|}{1 CSV} &
  \multicolumn{1}{c|}{1 CSV} &
  \multicolumn{1}{c|}{1 CSV} &
  \multicolumn{1}{c|}{2 CSV} &
  \multicolumn{1}{c|}{3 CSV} &
  \multicolumn{1}{c|}{2 CSV} &
  2 CSV &
  13 CSV &
  CODICE XML Entry \\
\end{tabular}%
}
\end{table}

\subsection{Methodology and Setup}

This section describes how BLINKG is used to evaluate the performance of the different LLMs presented in the previous section. It is important to note that our goal is not to compare different prompting strategies or output formats. The primary objective of BLINK is to provide a set of resources and a baseline evaluating general-purpose solutions. The intention is to establish a starting point for a new line of research focused on developing more effective and specialized approaches. The proposed procedure is designed to be generalizable across all benchmark scenarios.

We provide an overview of the benchmark size and configuration in Table~\ref{tab:experiments}, which reports, for each scenario, the number of instances per evaluation task, the ontology in use, and the characteristics of the input data. As we detailed in the benchmark description, Scenario~1 consists of eight progressively more complex cases, where we use CSV files as input data; we also tested the same experiments switching to JSON and XML, but we did not find significant variances. Scenario~2 relies on a complete version of the LinkedGTFS ontology and aggregates a total of 72 gold-standard mapping instances across the identified tasks. We use 13 GTFS-CSV files as input, provided as GTFS-1-CSV by GTFS-Madrid-Bench~\cite{chaves2020gtfs}. In order to stay within the LLMs’ context-window limits, we keep the full schema (i.e., column names) but only a small, representative sample of 5–10 rows per CSV in this scenario. Scenario~3 is grounded in a CODICE XML entry (see Listing~\ref{lst:ppds-entry}) mapped to a subset of ePO, with 26 task instances; here we use the full XML fragment, since data references are expressed as XPath expressions.

The evaluation process begins by constructing a standardized prompt template to ensure consistency across use cases. Each prompt clearly defines the task the LLM is expected to perform and includes a request to output the results in a structured table, with one column per predefined mapping task. This formatting constraint minimizes variability and facilitates direct comparison across models. The prompt also incorporates the data sources associated with the use case. For each source, its filename is included along with either its full content or, if input length becomes a constraint, a representative subset that preserves the original file’s structure and semantics. The ontology and SKOS taxonomies, defined in OWL and RDF respectively, are appended at the end of the prompt to ensure all necessary context is available to the model. This setup is designed to isolate model behavior from prompt formulation artifacts and provide traceable, interpretable outputs. We also evaluate different prompting strategies (zero-shot, one-shot, and few-shot) in the most complex setting (Scenario 3), and providing examples for the various tasks involved. The same prompt is run 3 times.

\begin{consultabox}
\textit{I have these/this input file(s) [that follow this schema [CODICE/GTFS]] and I want to build a knowledge graph with all possible mappings between them using this ontology. Could you make a table with the data references and a links to the corresponding ontology properties? Link all the properties that you can with the information that you have. Provide also the class of the entities or both the classes that they relate to, and a way to generate the subject of them. Use the following header: Data Reference, Ontology Property, Class Entity, ...

\{INPUT\_FILES\}

...

\{ONTOLOGY+SKOS\}

...
}
\end{consultabox}

To cope with the context-window limitations of the LLMs, we had to split Scenarios 2 and 3 into smaller sub-scenarios. Concretely, we isolated groups of ontology entities and built sub-scenarios around them, thereby reducing the number of classes, properties, and input files that needed to be processed in a single prompt. This allowed us to stay within the available token budget while preserving the structure of the original tasks. After running the models on each sub-scenario, we merged the outputs to compute the overall metrics for the full scenario. In practice, Scenario 2 was decomposed into six sub-scenarios, whereas Scenario 3 was split into two.


Before conducting the evaluation of the actual tasks, it is necessary to align each row in the model output with its corresponding row in the gold standard. To achieve this, we apply a two-step matching strategy. The primary approach is to use the ontology property column, which typically contains unique values per row. We compute similarity using the metrics described in Section 3.3 (i.e. Levenshtein distance and embedding-based similarity) to identify the closest match. If this initial matching fails (due to duplicate values or low similarity) we extend the matching criteria by incorporating two additional columns: ontology entity class and data reference. These columns were selected based on an analysis of multiple test cases from Scenario 1, where they consistently exhibited the highest similarity scores and helped disambiguate otherwise uncertain matches. The way of matching rows of tables is inspired by the baselines presented in Pugnaloni et al.~\cite{pugnaloni2025table}.

For each of the proposed scenarios, we conducted three different evaluation procedures. These are designed not only to measure the performance of the LLMs on the mapping tasks, but also to assess the effectiveness of the evaluation metrics introduced earlier:
\begin{itemize}
    \item \textbf{Expert evaluation}: Multiple experts in knowledge graph construction manually validate each LLM output to determine whether the identified mappings are correct with respect to the gold standard.
    \item \textbf{Raw evaluation:} The performance of each LLM is assessed using the proposed metrics directly on the raw outputs, without any modification or filtering.
    \item \textbf{Post-processed evaluation:} A set of heuristics is applied to clean the raw outputs. These included typical cleaning steps such as removing extra whitespace, discarding incoherent fragments, and eliminating incorrect columns from unrelated tasks. The goal is to improve the structural consistency of the output and provide a more realistic input to the evaluation metrics.
\end{itemize}

\subsection{Results}
For each scenario, we report the results in two stages. First, we present the similarity scores to analyze in depth how each LLM performs across the different mapping tasks. Then, we include three separate figures showing the F-score values, one for each evaluation setting (expert-based, raw, and post-processed). Additionally, each scenario is accompanied by a table in Appendix \ref{appendix1}, detailing the precision, recall, and F-score for every task, LLM, and evaluation type.

To select an appropriate similarity threshold for our proposed metric, we test several values and chose the one that produced results most closely aligned with the expert evaluation. We measured this alignment using Mean Absolute Error (MAE), which quantifies the average absolute difference between the automatically computed metric values and those obtained from expert judgments. In our case, the optimal threshold (minimizing MAE) was found to be 0.8.
\begin{figure}[!t]
    \centering
    \begin{subfigure}{0.49\textwidth}
        \includegraphics[width=\linewidth]{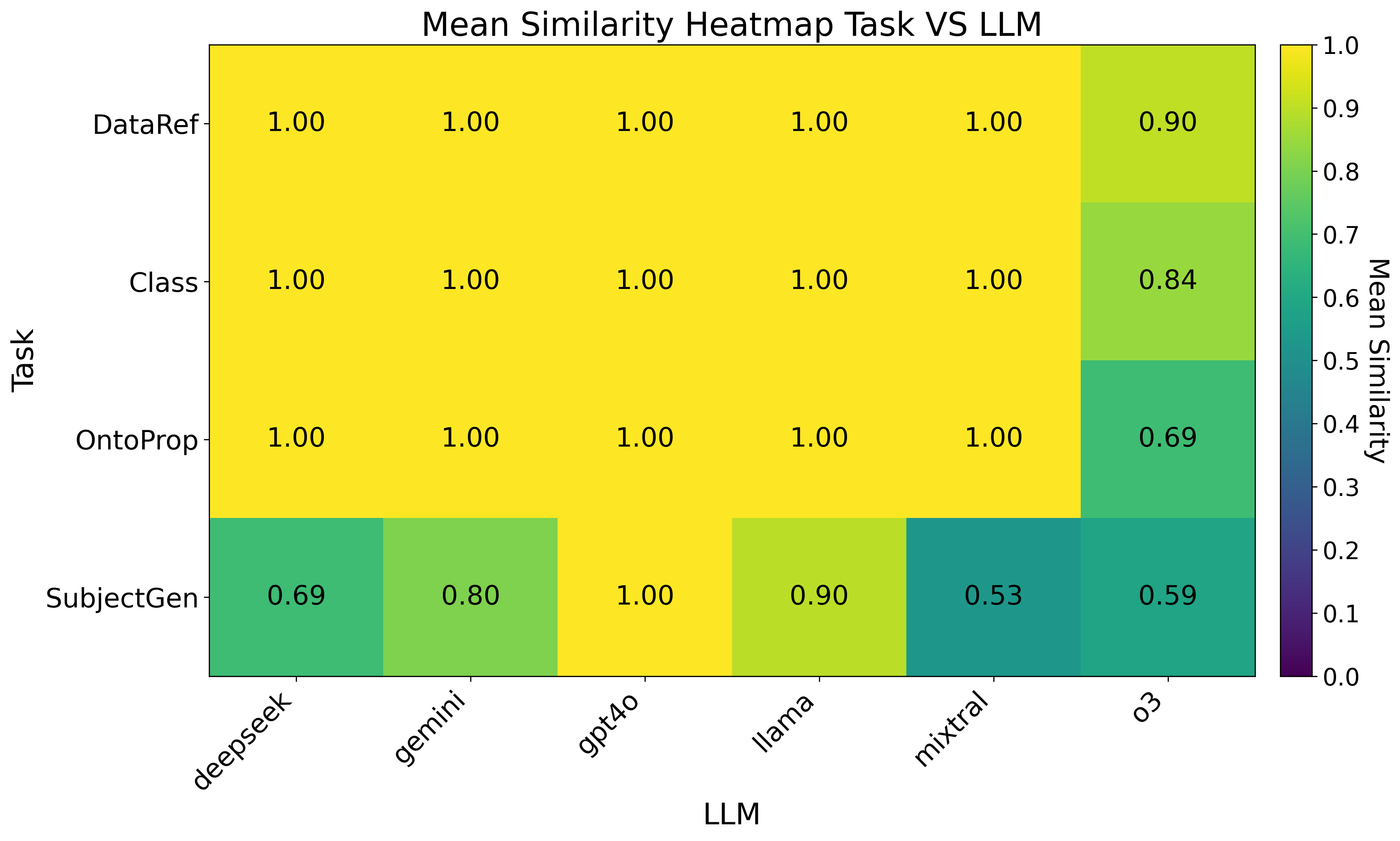}
        \caption{Similarity score for 1A}
    \end{subfigure}
    \begin{subfigure}{0.49\textwidth}
        \includegraphics[width=\linewidth]{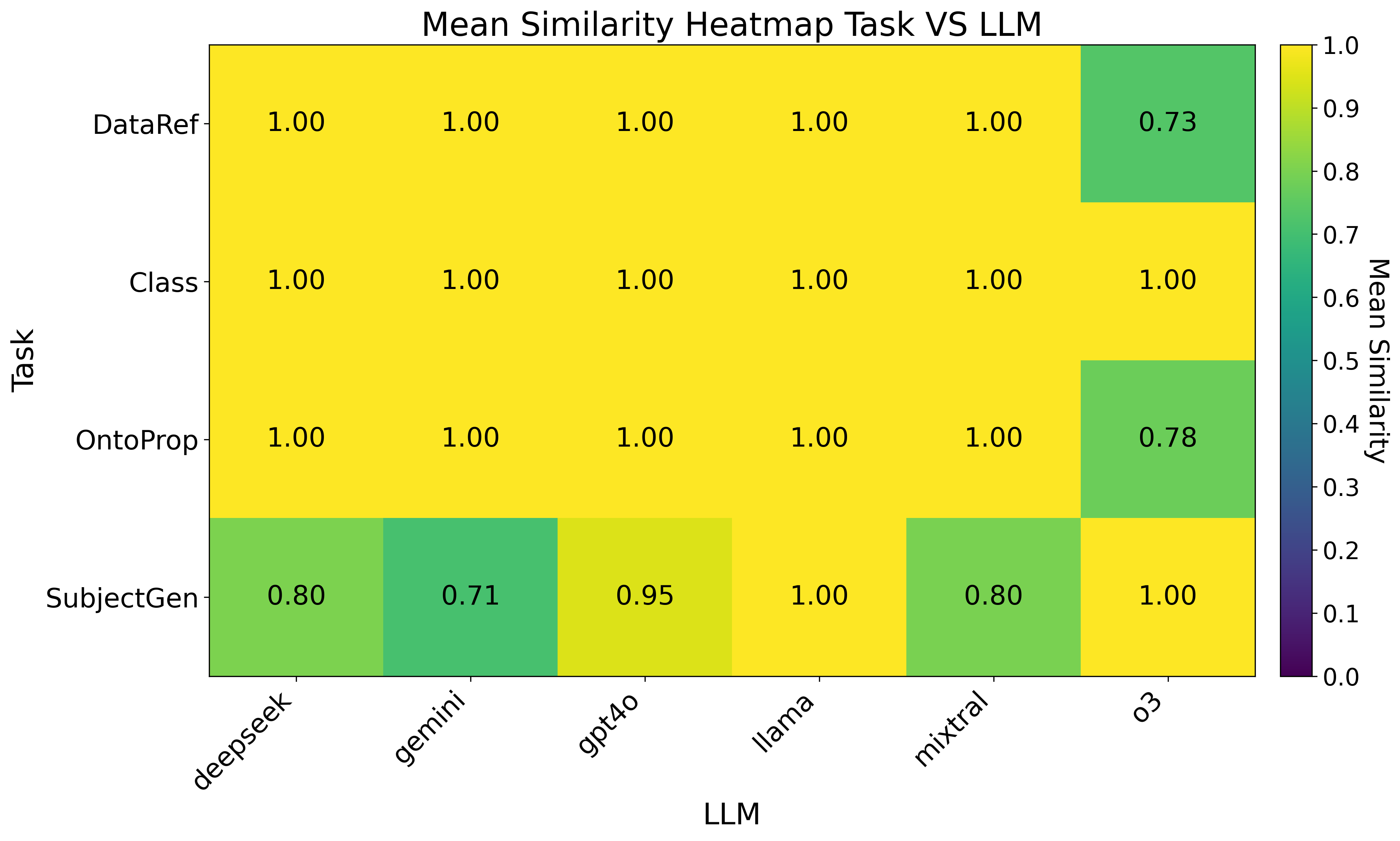}
        \caption{Similarity score for 1B}
    \end{subfigure}
    
    \vspace{0.5em}
    
    \begin{subfigure}{0.49\textwidth}
        \includegraphics[width=\linewidth]{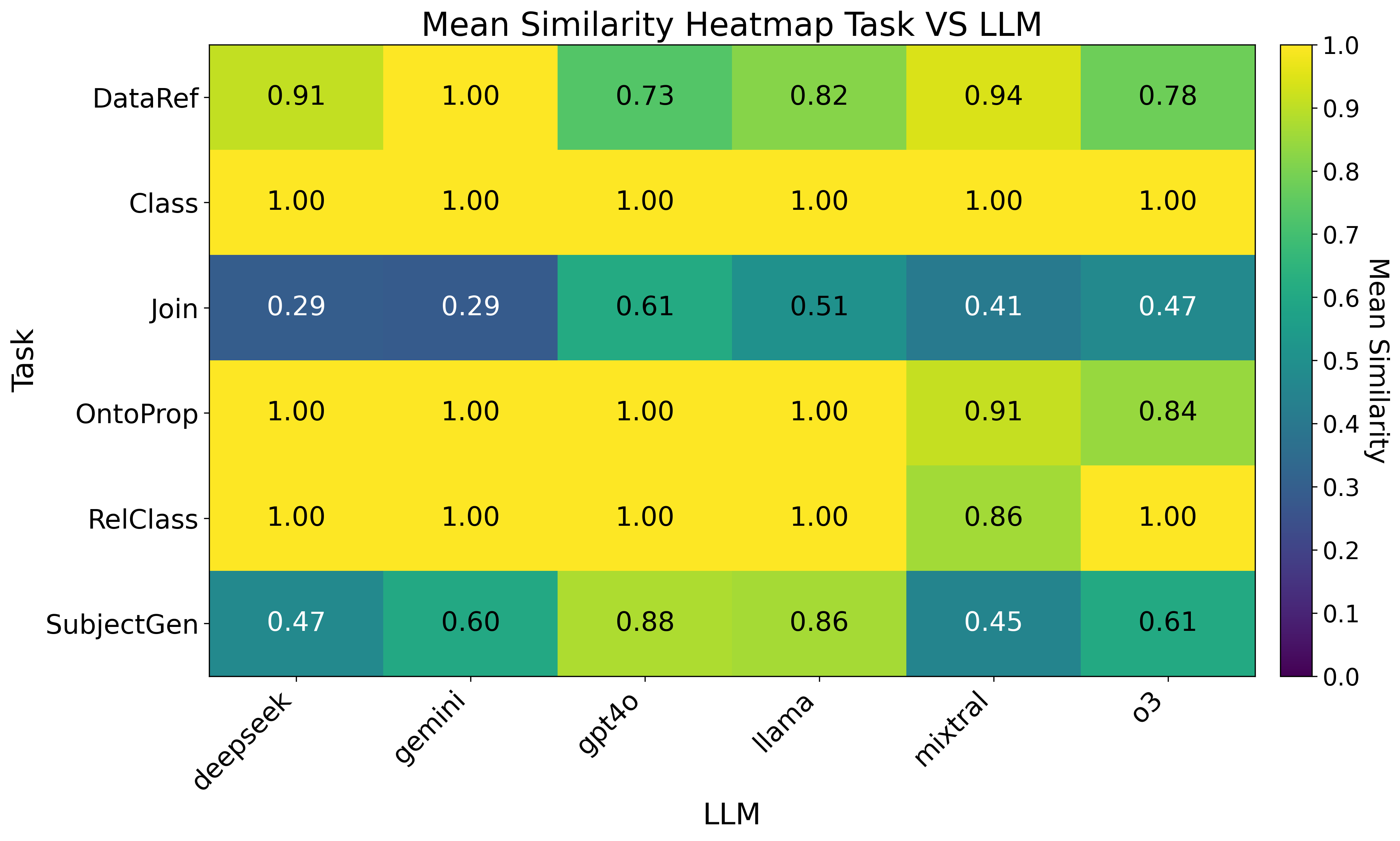}
        \caption{Similarity score for 1C}
    \end{subfigure}
    \begin{subfigure}{0.49\textwidth}
        \includegraphics[width=\linewidth]{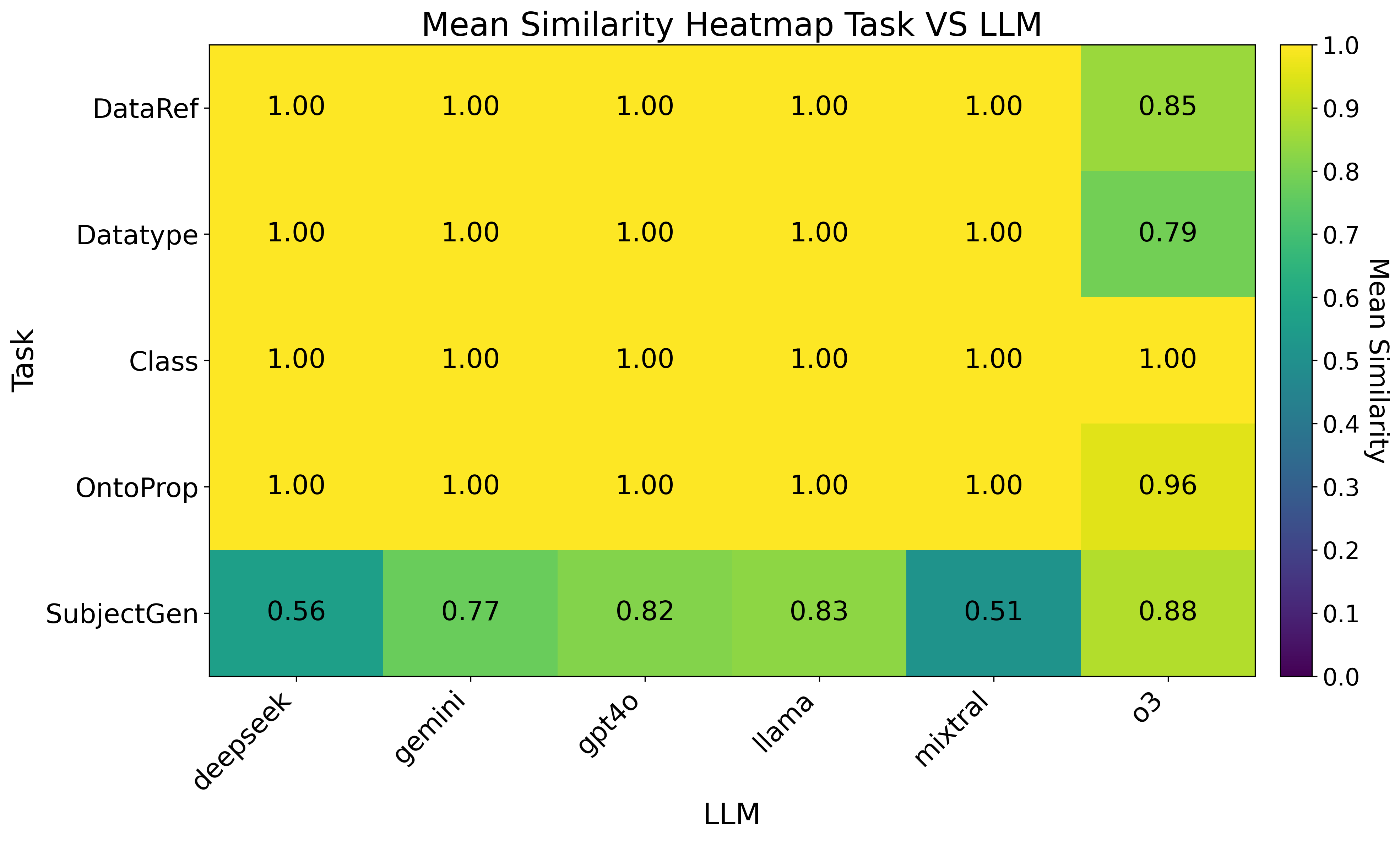}
        \caption{Similarity score for 1D}
    \end{subfigure}
    
    \vspace{0.5em}
    
    \begin{subfigure}{0.49\textwidth}
        \includegraphics[width=\linewidth]{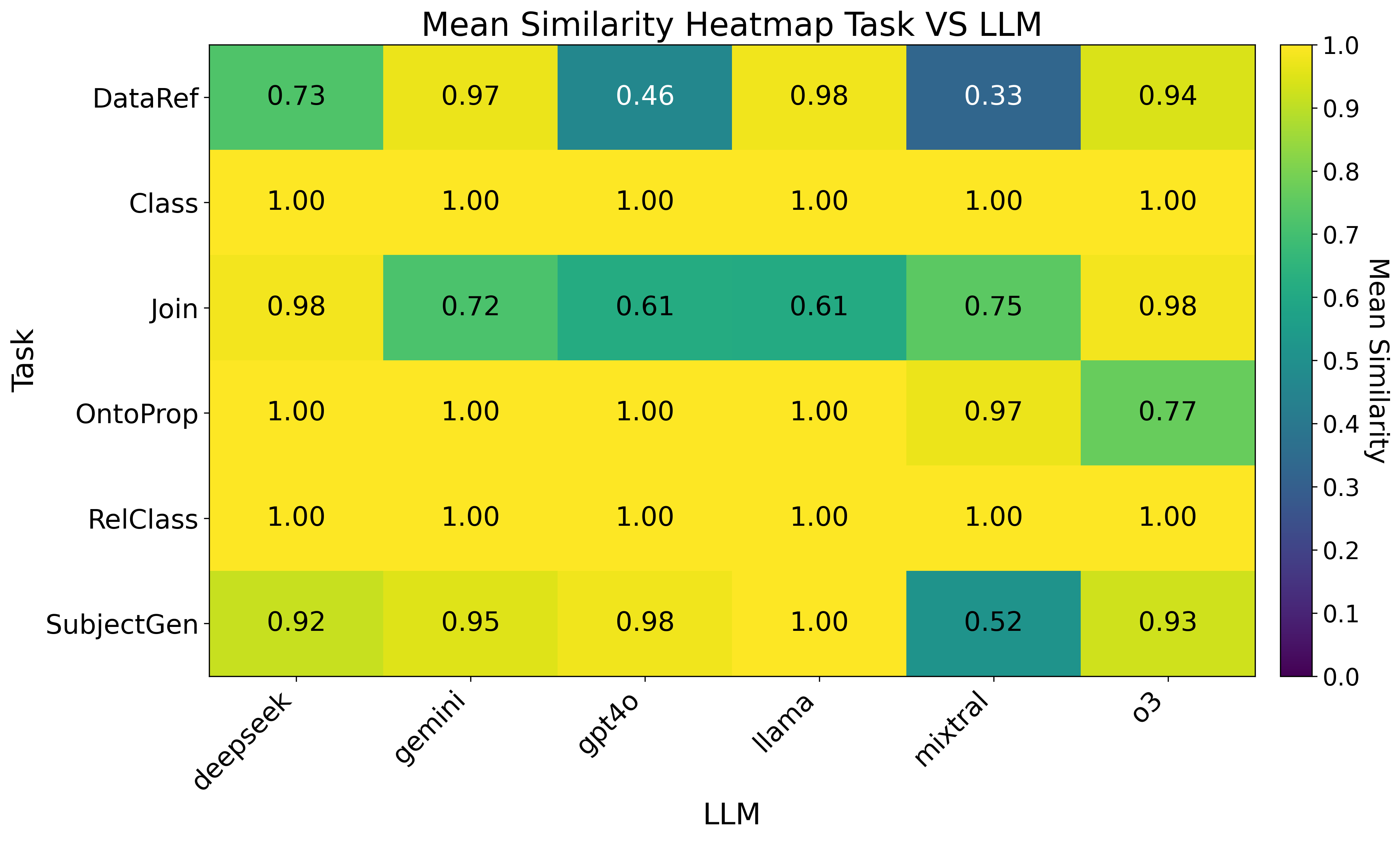}
        \caption{Similarity score for 1E}
    \end{subfigure}
    \begin{subfigure}{0.49\textwidth}
        \includegraphics[width=\linewidth]{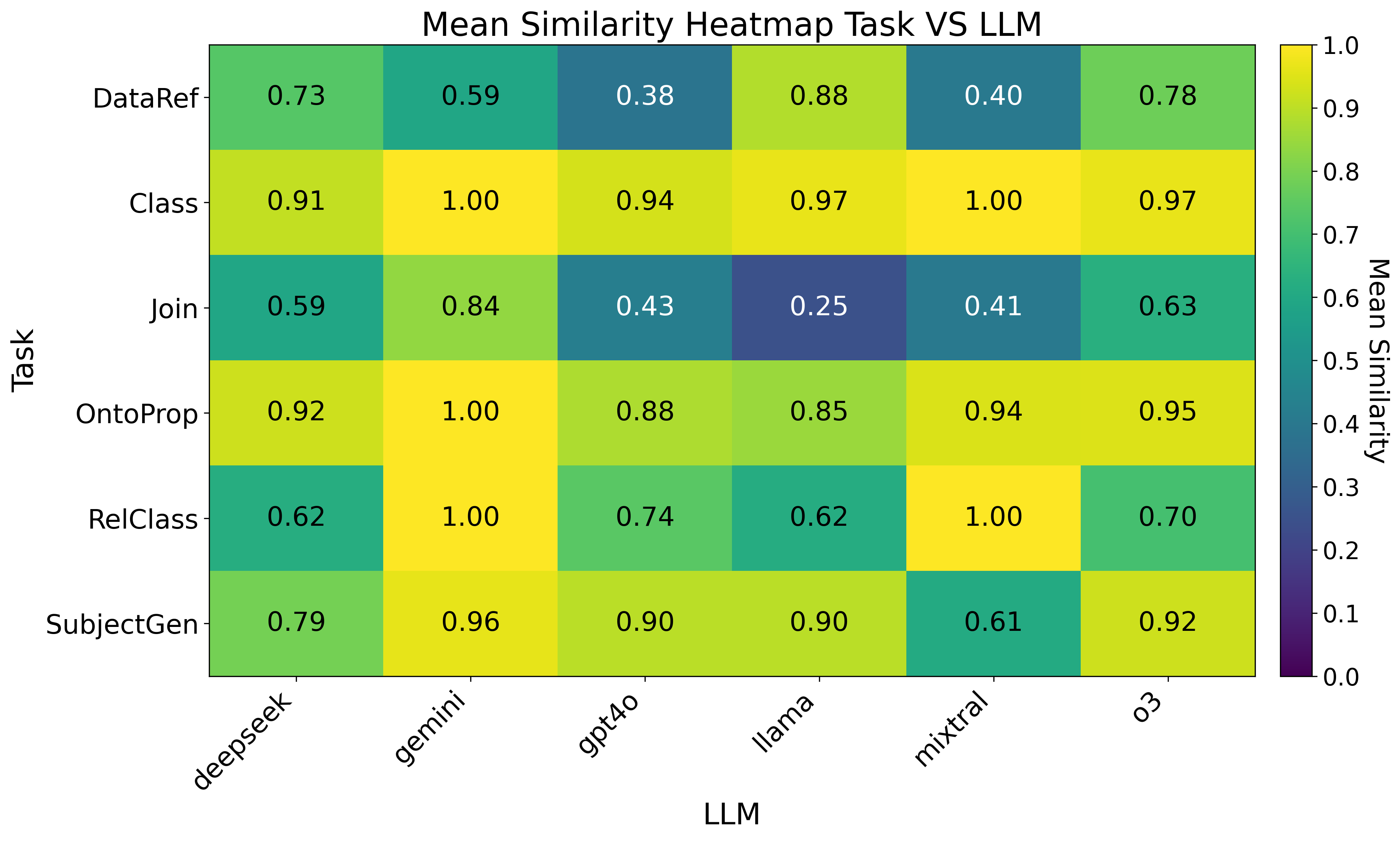}
        \caption{Similarity score for 1F}
    \end{subfigure}
    
    \vspace{0.5em}
    
    \begin{subfigure}{0.49\textwidth}
        \includegraphics[width=\linewidth]{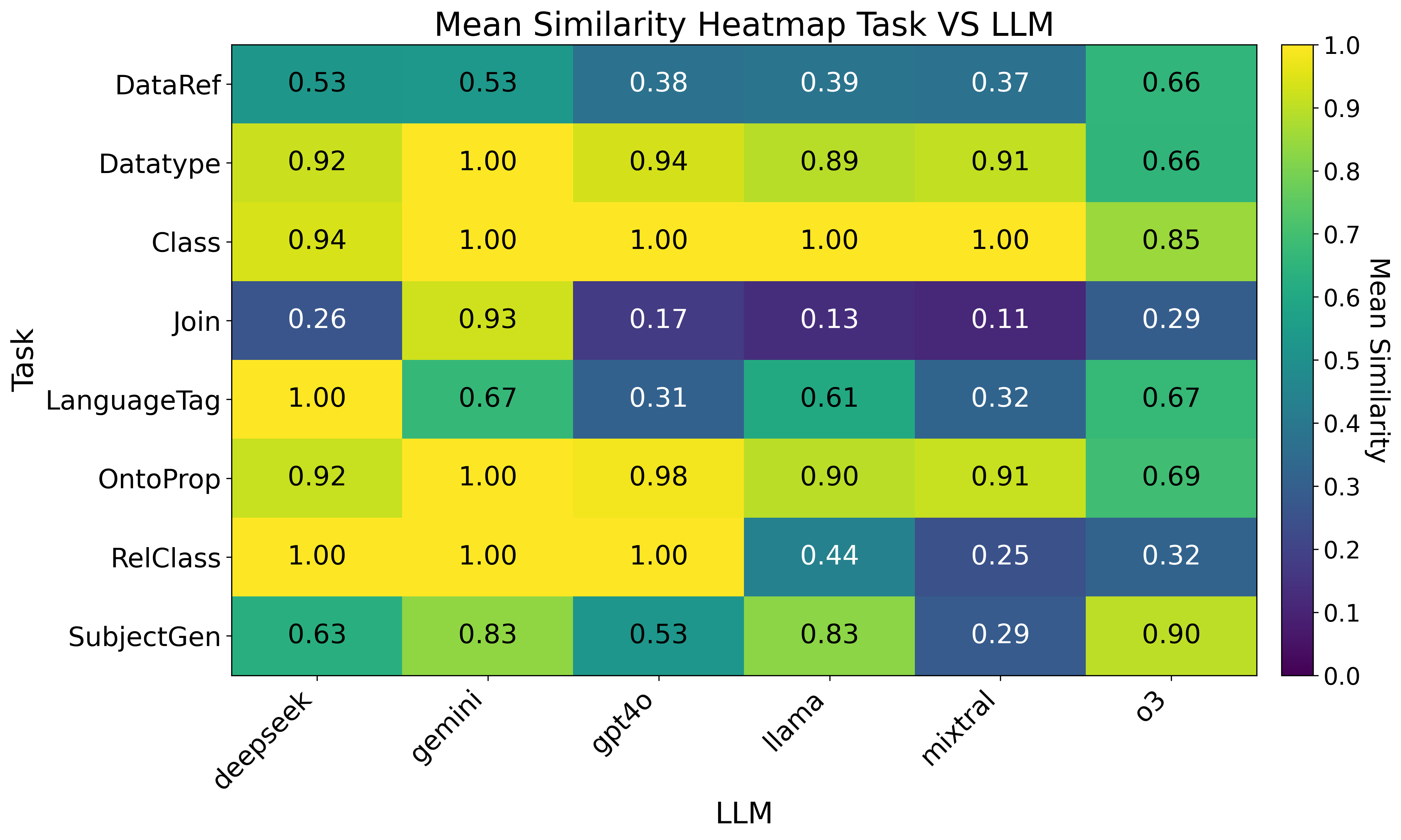}
        \caption{Similarity score for 1G}
    \end{subfigure}
    \begin{subfigure}{0.49\textwidth}
        \includegraphics[width=\linewidth]{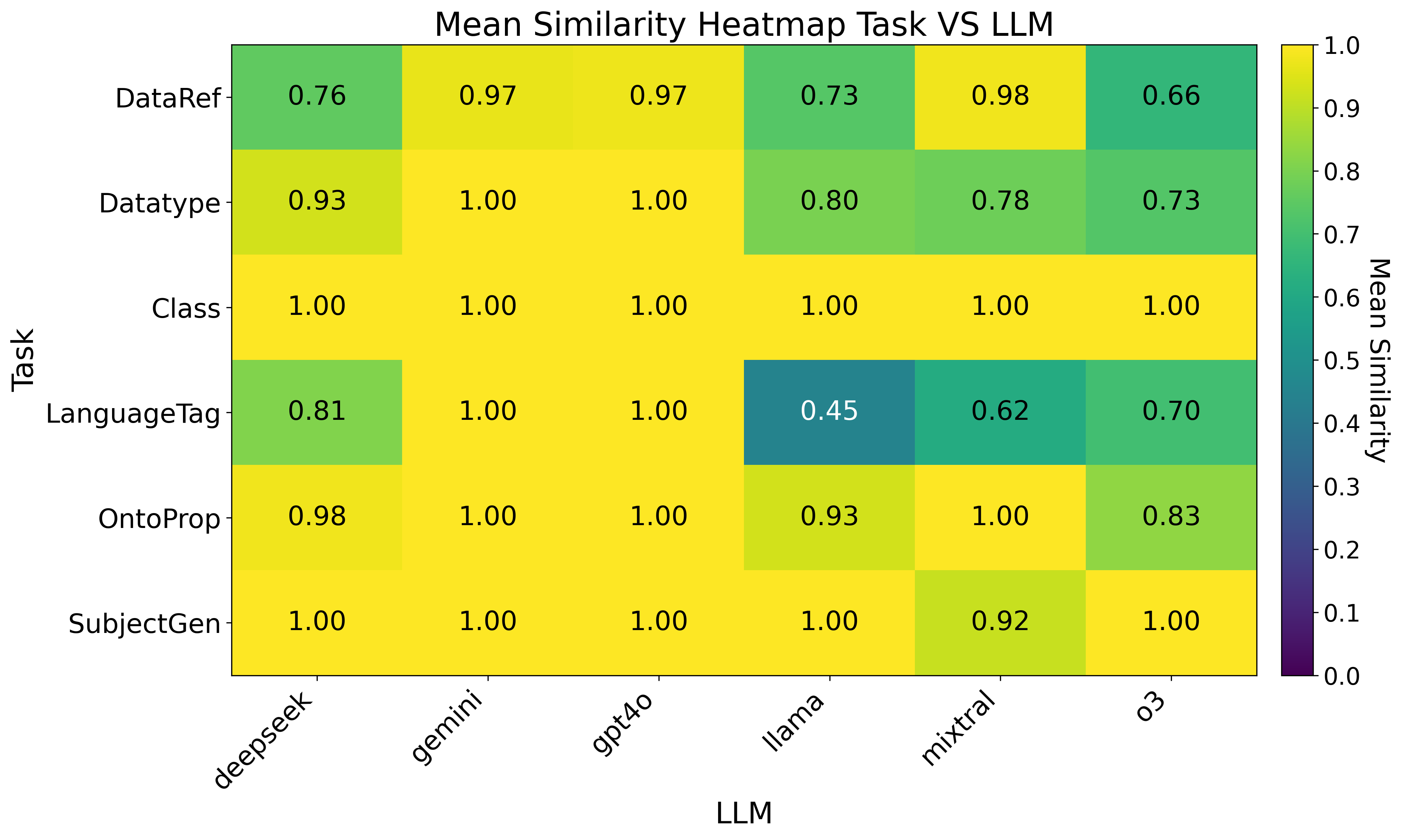}
        \caption{Similarity score for 1H}
    \end{subfigure}

    \caption{Similarity scores across the eight configurations with respect to the gold standard in the Scenario 1.}
    \label{fig:similarity_scenario1}
\end{figure}
\subsubsection*{Scenario 1}
\begin{figure}[!t]
\centering
    \begin{subfigure}{0.49\textwidth}
        \includegraphics[width=\textwidth]{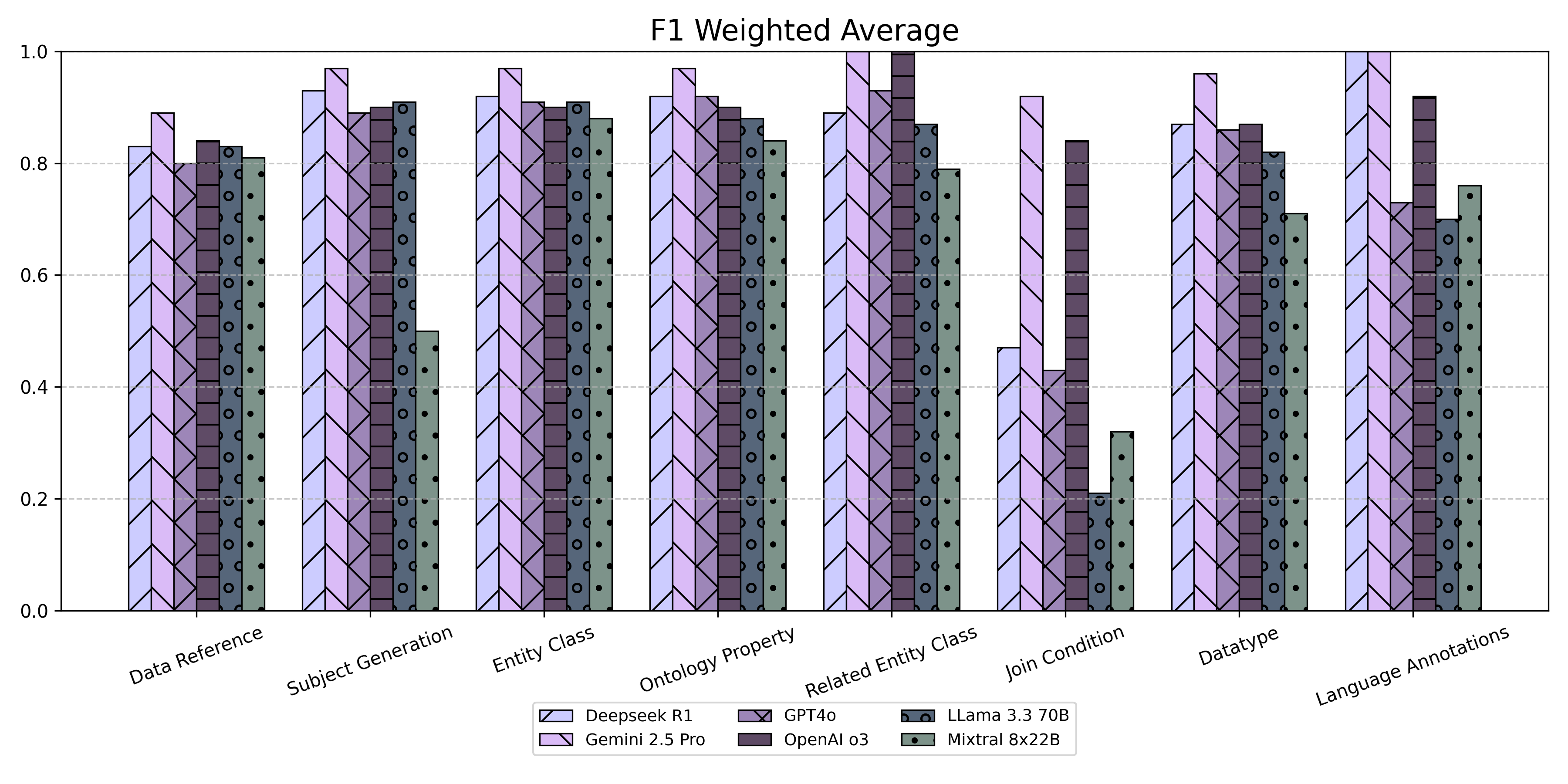}
    \caption{Expert F-score of the different tasks and LLMs.}
    \label{fig:results1}
    \end{subfigure}
    \hfill
    \begin{subfigure}{0.49\textwidth}
        \includegraphics[width=\textwidth]{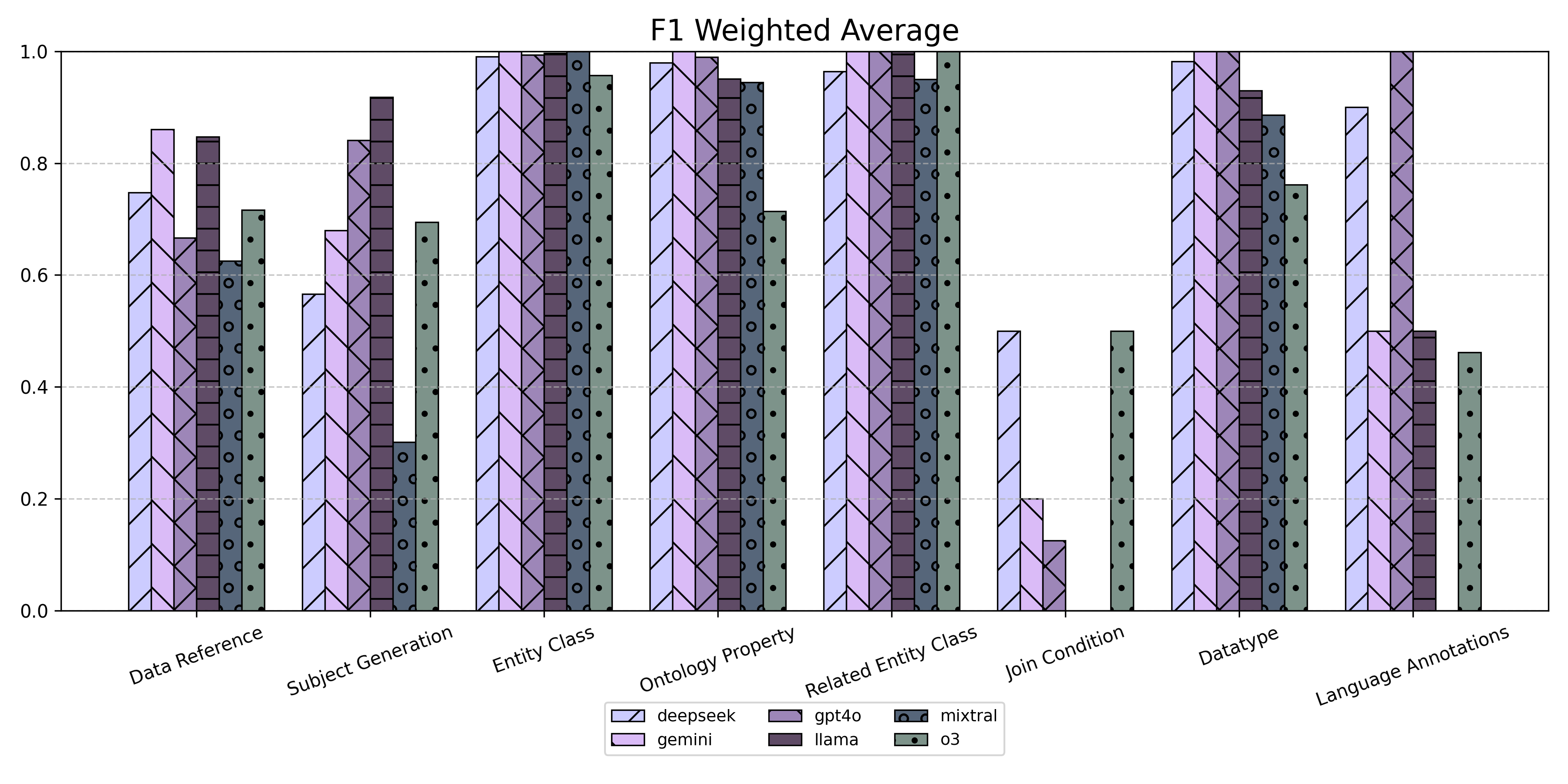}
    \caption{Raw F-score of the different tasks and LLMs.}
    \label{fig:results1_auto_noclean}
    \end{subfigure}
    \begin{subfigure}{0.49\textwidth}
        \includegraphics[width=\textwidth]{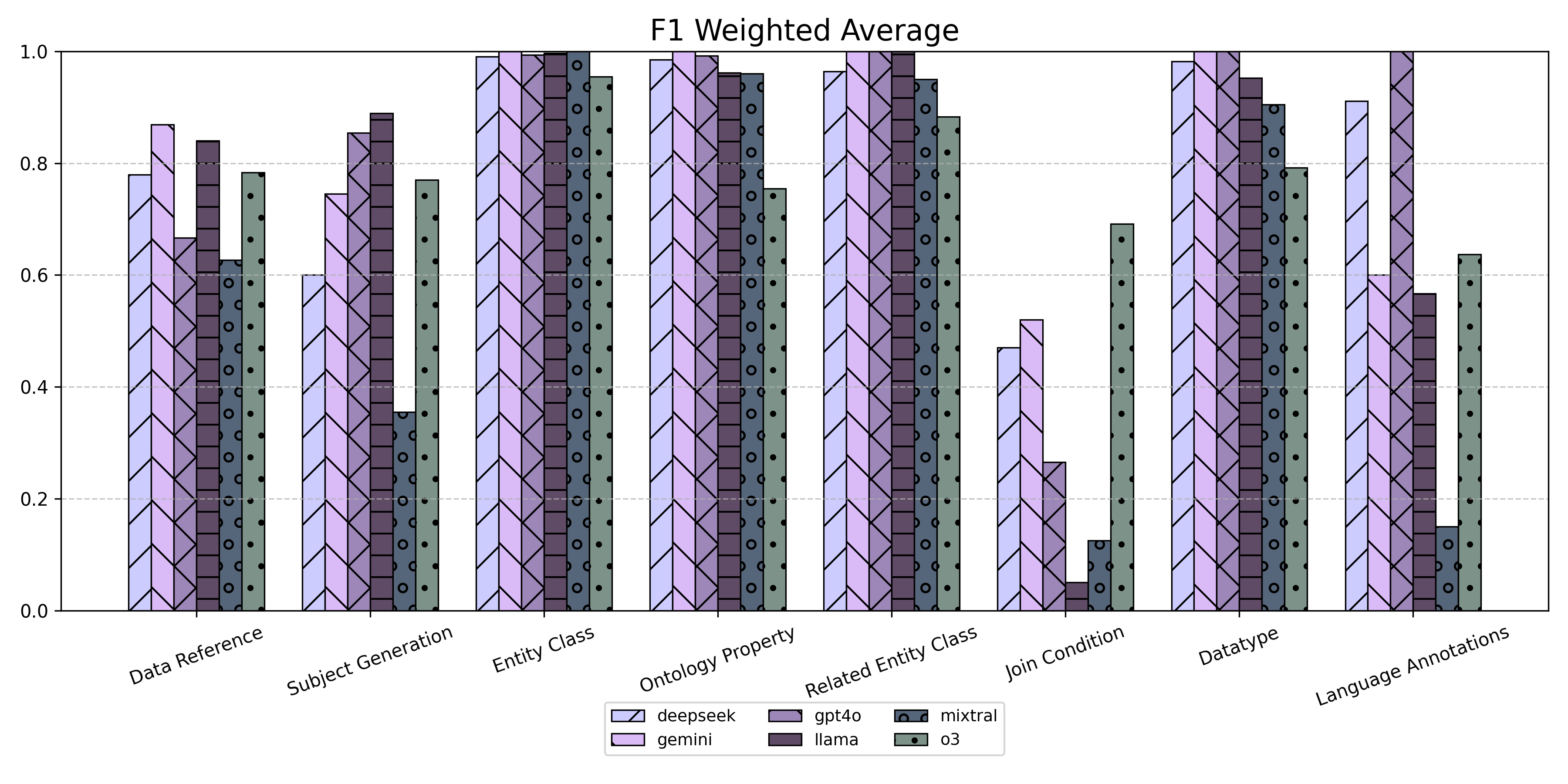}
    \caption{Post-processed F-score of the different tasks and LLMs.}
    \label{fig:results1_auto}
    \end{subfigure}
    \caption{Comparison of the F-score between the three different evaluations in the Scenario 1.}
    \label{fig:f1_sce1}
\end{figure}

Figure \ref{fig:similarity_scenario1} summarizes the average similarity scores across all atomic test cases in Scenario 1, grouped by task and LLM. We only provide the similarity over the post-processed outputs as there are no significant differences between them and the raw ones (see Figure \ref{fig:f1_sce1}) The results show that Entity Class identification consistently yields the highest similarity, with most models achieving values above 0.95, and several reaching perfect alignment with the gold standard. This suggests that LLMs are particularly effective at inferring the class of an entity from tabular data. Ontology Property identification also performs reliably across models, though with slightly more variability than entity classification. Scores remain high overall, indicating that models are able to associate columns with ontology properties in a consistent manner. Related Entity Class shows performance similar to the class detection, though slightly more sensitive to model differences.

Datatype prediction is generally strong for most models ($\geq 0.89$), with the exception of o3, which shows a lower similarity score (0.72), possibly due to inconsistencies in literal typing or formatting in the outputs. In contrast, Join Condition generation remains the most challenging task, with similarity scores ranging from 0.37 (Llama) to 0.69 (Gemini 2.5). This reflects the difficulty LLMs face when inferring logical conditions to connect entities across data sources. Language Annotation exhibits substantial variance between models. Deepseek R1 and Gemini achieve relatively high scores (> 0.83), but models like Mixtral and LLama perform below 0.53. This inconsistency suggests that detecting or generating correct language tags is not yet robustly handled by current LLMs. Finally, Data Reference matching shows medium-to-high performance across models. Gemini leads with an average score of 0.88, while GPT4o trails behind at 0.74.

As shown in Figure \ref{fig:f1_sce1}, the obtained F-score in the raw evaluation tends to underestimate model performance, particularly in tasks that are sensitive to formatting or structural correctness, such as Join Condition and Language Annotations. In environments as simple as those proposed in Scenario 1, the differences between raw and post-processed outputs are negligible. This indicates that LLMs are capable of closely adhering to the prompt instructions, producing structured outputs with minimal syntactic or formatting noise. As a result, postprocessing has little impact, since most of the relevant information is already correctly and cleanly expressed in the raw outputs. 

\subsubsection*{Scenario 2}

\begin{figure}[!t]
    \centering
    \begin{subfigure}{0.49\textwidth}
        \includegraphics[width=\textwidth]{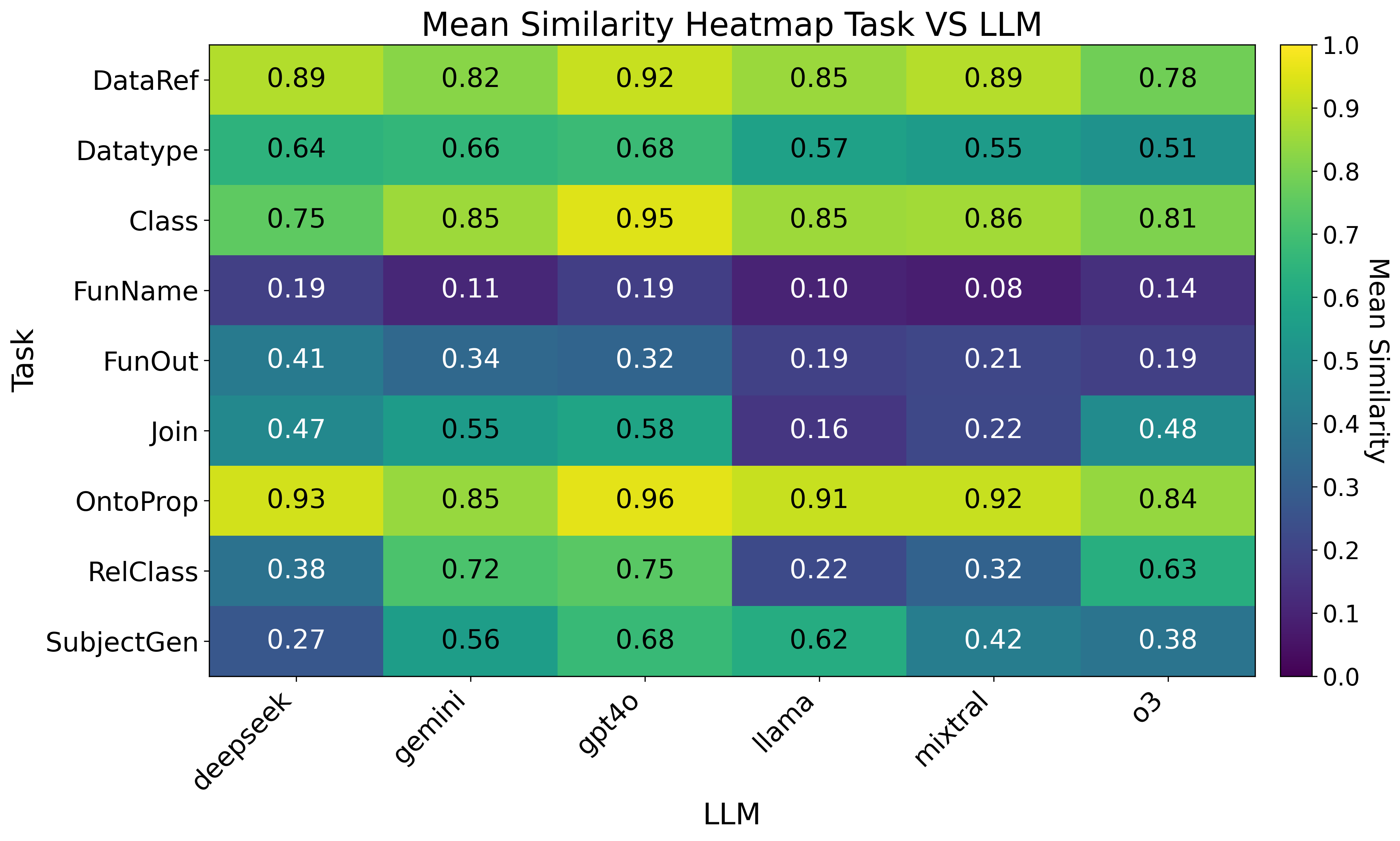}
        \caption{Similarity Raw Results}
    \end{subfigure}
    \begin{subfigure}{0.49\textwidth}
        \includegraphics[width=\linewidth]{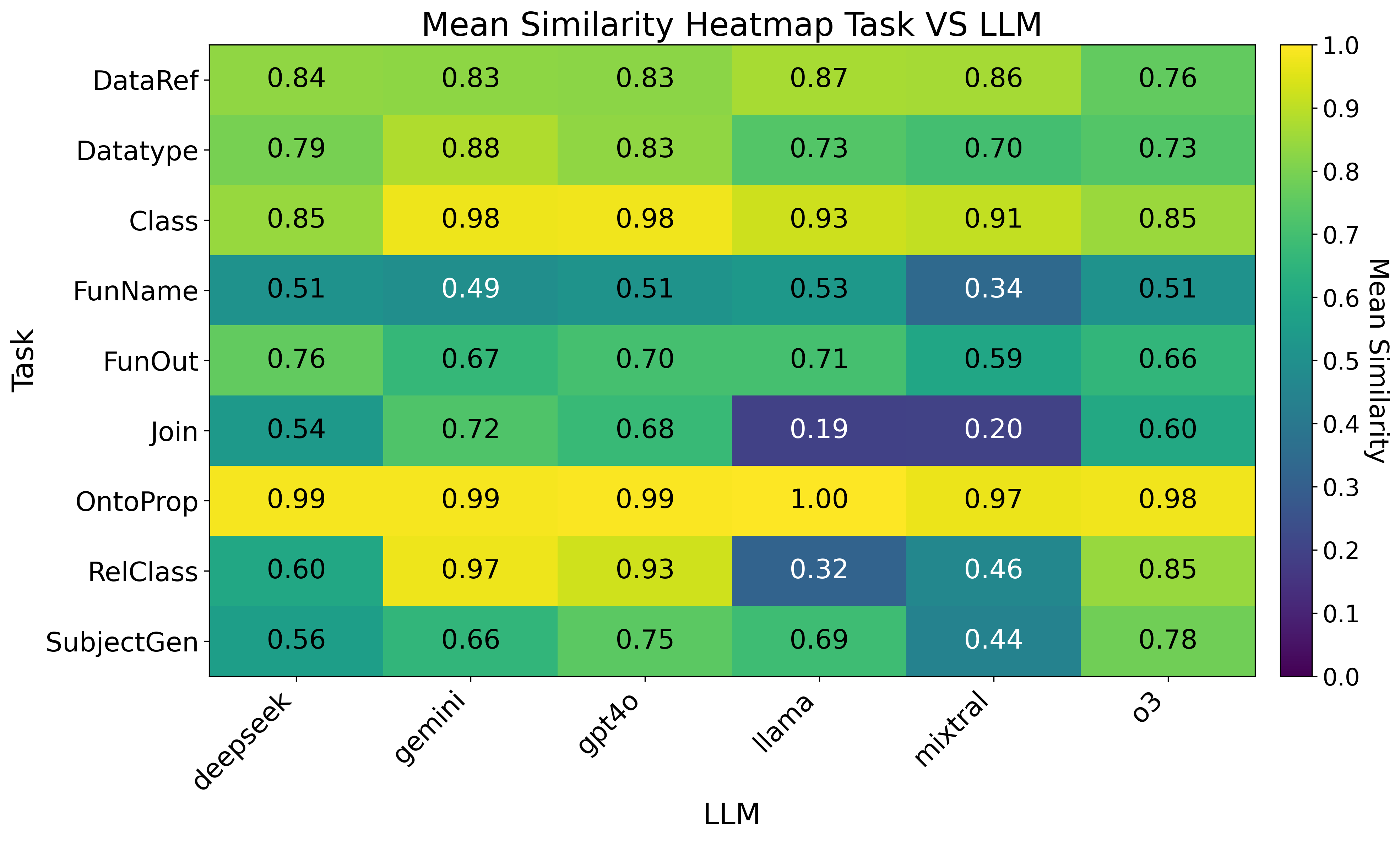}
        \caption{Similarity Post-processed Results}
    \end{subfigure}
    \caption{Comparison of similarity metrics between raw and post-processed configurations with respect to the gold standard in Scenario 2.}
    \label{fig:similarity_scenario2}
\end{figure}

Scenario 2 was designed to test LLMs in tasks that go beyond simple entity or property alignment, requiring them to handle structured data transformations and schema-level logic. As shown in the similarity results (Figure~\ref{fig:similarity_scenario2}), this scenario exposes greater variability across LLMs and highlights significant gaps between raw and post-processed outputs. \textit{Entity Class} and \textit{Data Reference} remain the most reliably handled tasks, with post-processed similarity scores exceeding 0.90 in several models—most notably GPT4o (0.91 in \textit{Data Reference}) and Deepseek (0.95 in \textit{Entity Class}). These tasks are typically guided by strong lexical cues in the input and are less affected by verbosity or syntactic inconsistencies.

Other concept-matching tasks, such as \textit{Ontology Property} and \textit{Related Entity Class}, also achieve strong performance. For example, Deepseek improves from 0.74 (raw) to 0.80 (post-processed) in \textit{Ontology Property}, while Gemini reaches 0.86 in \textit{Related Entity Class}. This suggests that LLMs can effectively leverage contextual signals to identify relevant ontology terms, as long as surface-level alignment exists and the output is properly cleaned.

In contrast, tasks requiring structural reasoning show much weaker results. \textit{Join Condition}, in particular, remains challenging for all models: even after post-processing, no model exceeds 0.60. The difficulty in inferring relational logic between entities mirrors observations from Scenario 1 and illustrates the persistent gap between language generation capabilities and symbolic reasoning.

Function-related tasks perform worst. Most models score below 0.20 in \textit{Function Name} and below 0.41 in \textit{Function Output}, even after cleaning. These tasks require abstraction, domain-specific function knowledge, and precise formatting—areas where current LLMs often hallucinate, misinterpret, or produce vague and irrelevant output. Similarly, \textit{Subject Generation} remains sensitive to output formatting, but shows moderate improvement with post-processing (e.g., GPT4o rises from 0.64 to 0.71), indicating that some errors are more syntactic than semantic.

\begin{figure}[t]
\centering
    \begin{subfigure}{0.49\textwidth}
        \includegraphics[width=\textwidth]{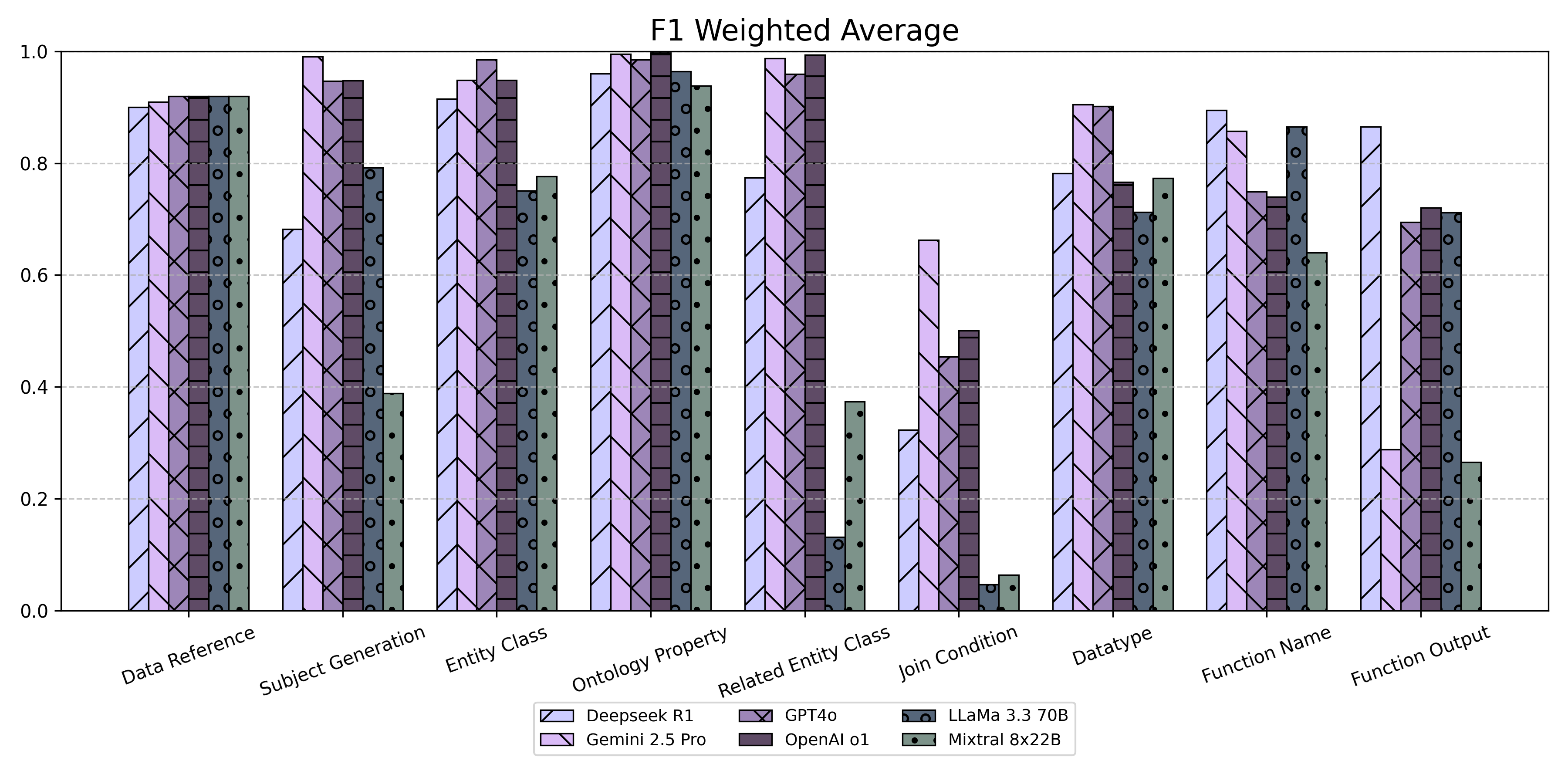}
    \caption{Expert F-score of the different tasks and LLMs.}
    \label{fig:results2}
    \end{subfigure}
    \hfill
    \begin{subfigure}{0.49\textwidth}
        \includegraphics[width=\textwidth]{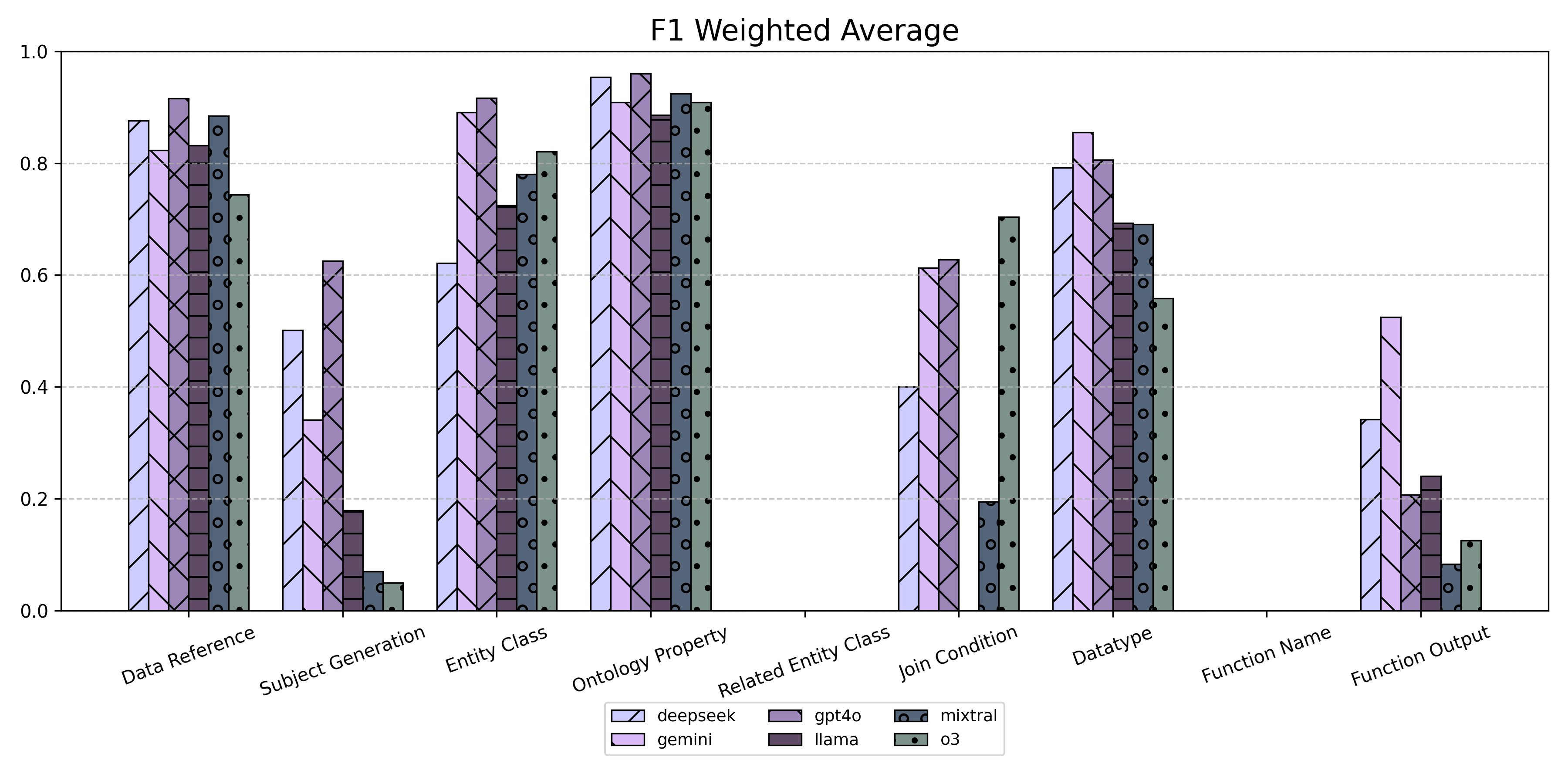}
    \caption{Raw F-score of the different tasks and LLMs.}
    \label{fig:results2_auto_noclean}
    \end{subfigure}
    \begin{subfigure}{0.49\textwidth}
        \includegraphics[width=\textwidth]{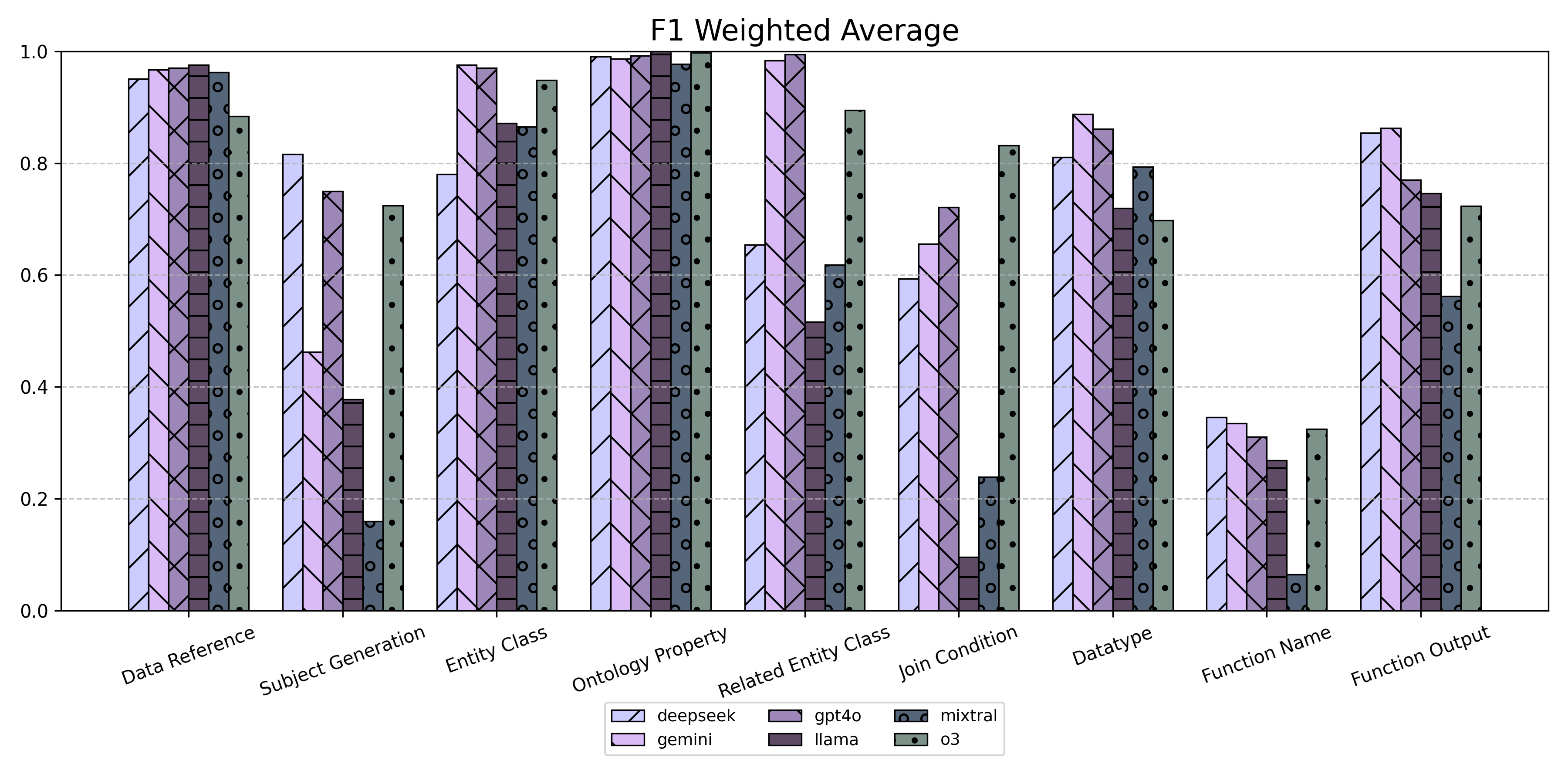}
    \caption{Post-processed F-score of the different tasks and LLMs.}
    \label{fig:results2_auto}
    \end{subfigure}
    \caption{Comparison of the F-score between the three different evaluations in the Scenario 2.}
    \label{fig:f1_sce2}
\end{figure}

F-score scores (Figure~\ref{fig:f1_sce2}) align with these observations. Manual annotations consistently yield high results across well-structured tasks, with several models exceeding 0.94 in \textit{Entity Class} and \textit{Subject Generation}. However, raw evaluations often underestimate LLM performance due to noisy output structures. For instance, in \textit{Join Condition}, some raw scores drop by over 0.30 compared to the manual gold standard. Post-processing significantly narrows this gap in most tasks, particularly for \textit{Ontology Property} and \textit{Function Output}, where models like GPT4o and Deepseek improve by 20+ points after cleaning. Nevertheless, difficult tasks like \textit{Function Name} and \textit{Join Condition} still show marked discrepancies, highlighting the limitations of current LLMs and automatic scoring.

Overall, Scenario 2 confirms that while LLMs are increasingly capable in aligning entities and interpreting schema-level concepts when lexical clues are present, they continue to struggle with structural logic and semantic transformation. Post-processing is essential to extract structured meaning from raw outputs, but human-in-the-loop evaluation remains critical to ensure correctness in more abstract or structurally complex mapping tasks.

\subsubsection*{Scenario 3}

\begin{figure}[!t]
    \centering
    \begin{subfigure}{0.49\textwidth}
        \includegraphics[width=\linewidth]{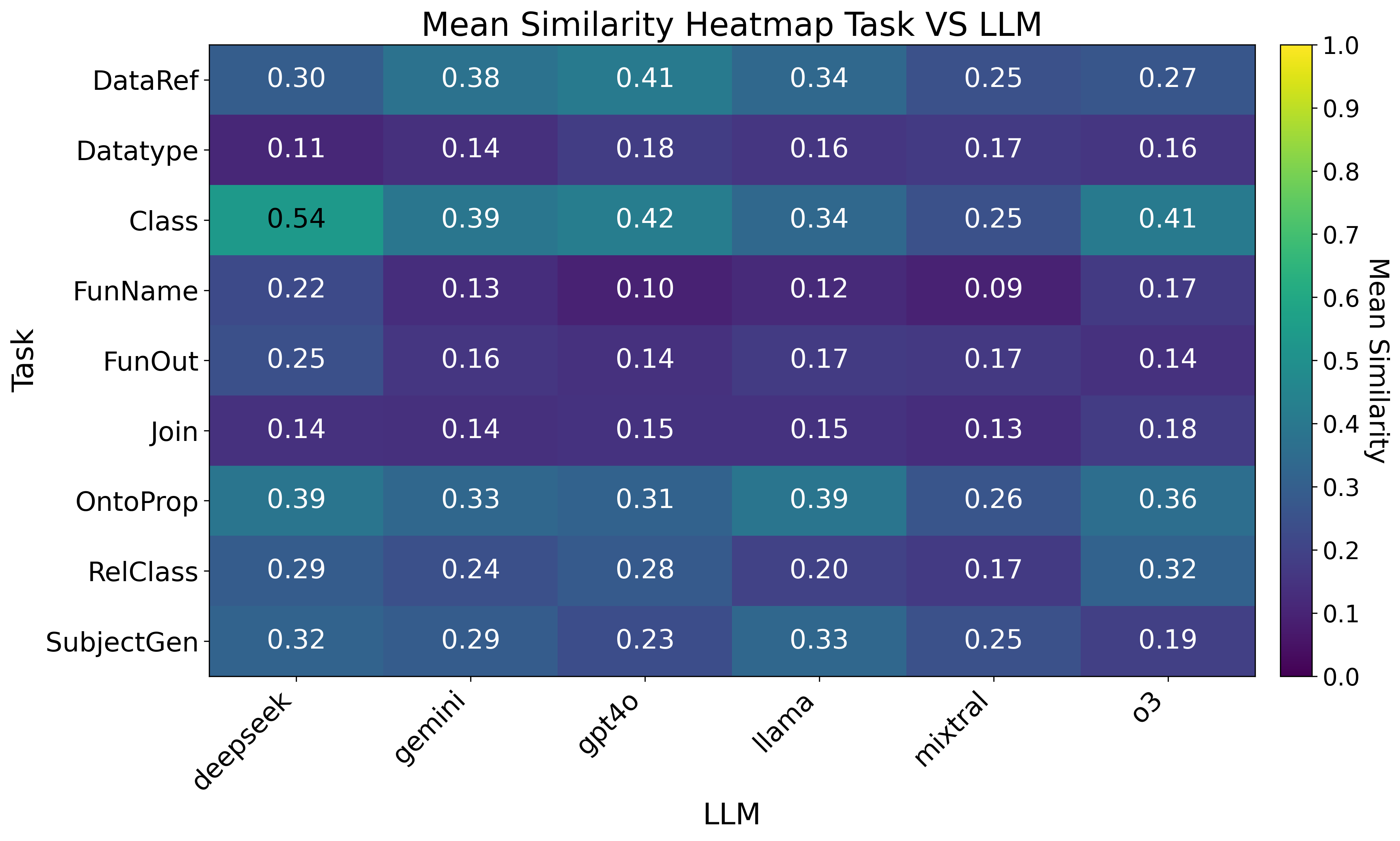}
        \caption{Similarity Raw Results}
    \end{subfigure}
    \hfill
    \begin{subfigure}{0.49\textwidth}
        \includegraphics[width=\linewidth]{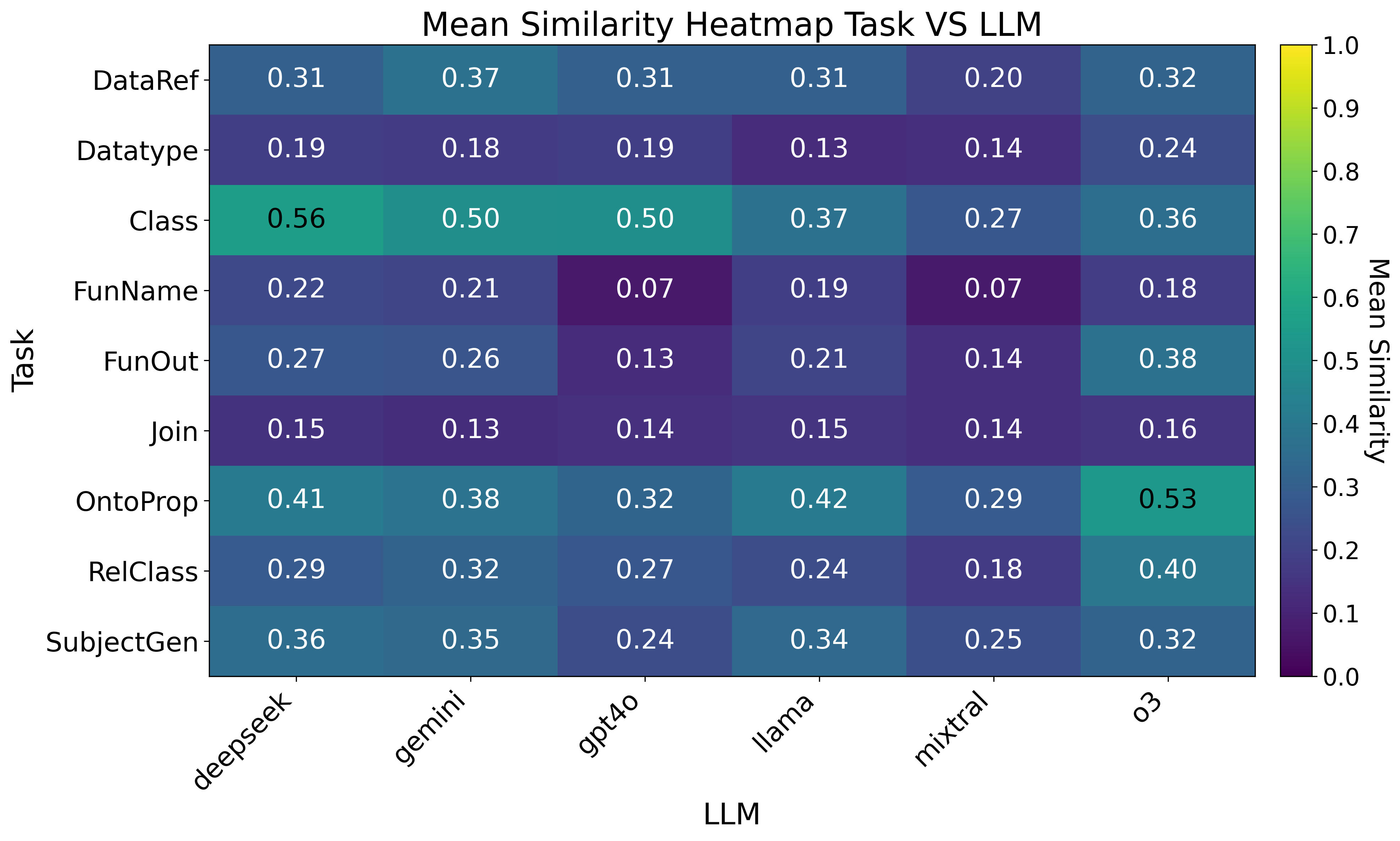}
        \caption{Similarity Post-processed Results}
    \end{subfigure}

    \caption{Comparison of similarity metrics between raw and post-processed configurations with respect to the gold standard in Scenario 3.}
    \label{fig:similarity_scenario3}
\end{figure}

Scenario 3 was designed to challenge LLMs with schema-distant inputs, where the structure and terminology of the source data differ significantly from the target ontology. As shown in the similarity results (Figure~\ref{fig:similarity_scenario3}), this setting reveals critical limitations in the models’ ability to generalize and abstract. Across all tasks, we observe a sharp drop in similarity scores compared to Scenarios 1 and 2, with no task surpassing a mean of 0.53—even after postprocessing.

Among all tasks, \textit{Entity Class} remains the best-performing one, with Deepseek achieving 0.53 and GPT4o 0.42. Other concept alignment tasks, such as \textit{Related Entity Class} and \textit{Ontology Property}, yield moderate results (0.35–0.47), but fall significantly short of earlier scenarios. For example, Gemini reaches 0.47 in \textit{Ontology Property}, yet fails to generalize in structurally dependent tasks. This suggests that while some lexical cues may still be leveraged, LLMs struggle to ground predictions in ontology-aware semantics without alignment.

Performance deteriorates further in structurally complex tasks. \textit{Join Condition}, \textit{Datatype}, and \textit{Subject Generation} typically remain below 0.40, even for strong models like GPT4o. Function-related tasks (\textit{Function Name} and \textit{Function Output}) are especially problematic, with all models scoring below 0.25. The models often produce verbose, unrelated operations, failing to extract functional meaning from schema-distant representations.

\begin{figure}[!t]
\centering
    \begin{subfigure}{0.49\textwidth}
        \includegraphics[width=\textwidth]{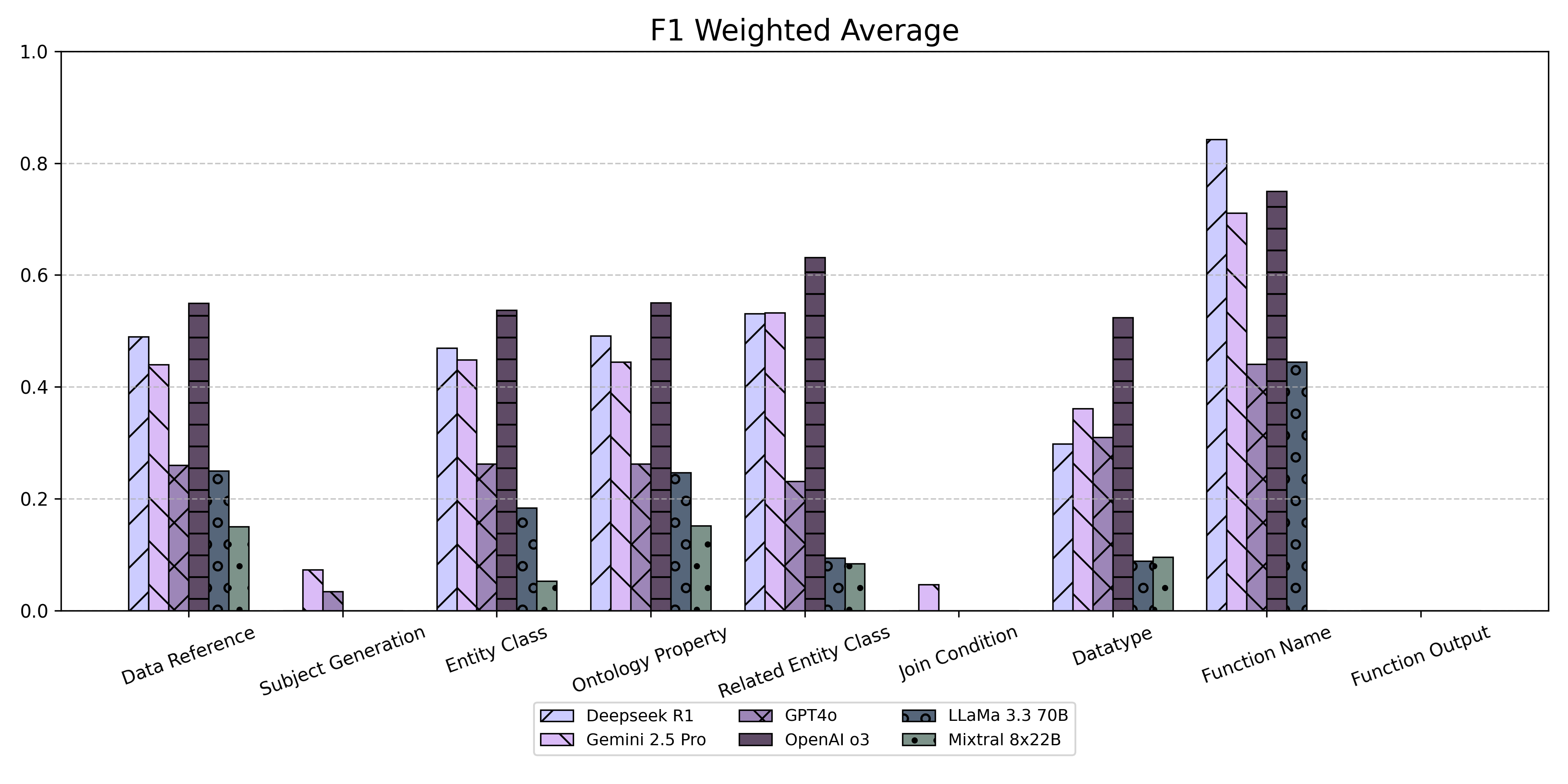}
    \caption{Expert F-score of the different tasks and LLMs.}
    \label{fig:results3}
    \end{subfigure}
    \hfill
    \begin{subfigure}{0.49\textwidth}
        \includegraphics[width=\textwidth]{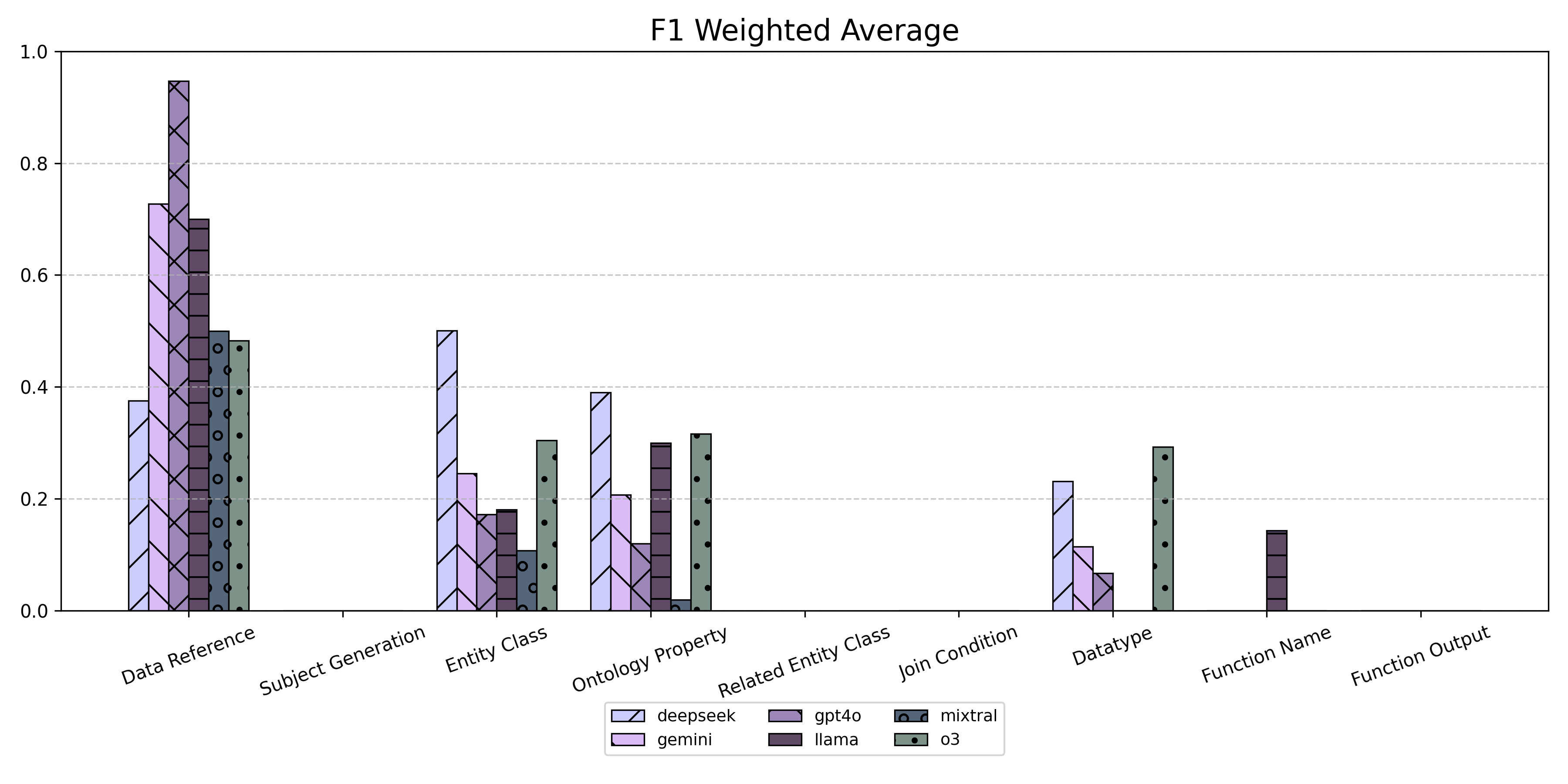}
    \caption{Raw F-score of the different tasks and LLMs.}
    \label{fig:results3_auto_noclean}
    \end{subfigure}
    \begin{subfigure}{0.49\textwidth}
        \includegraphics[width=\textwidth]{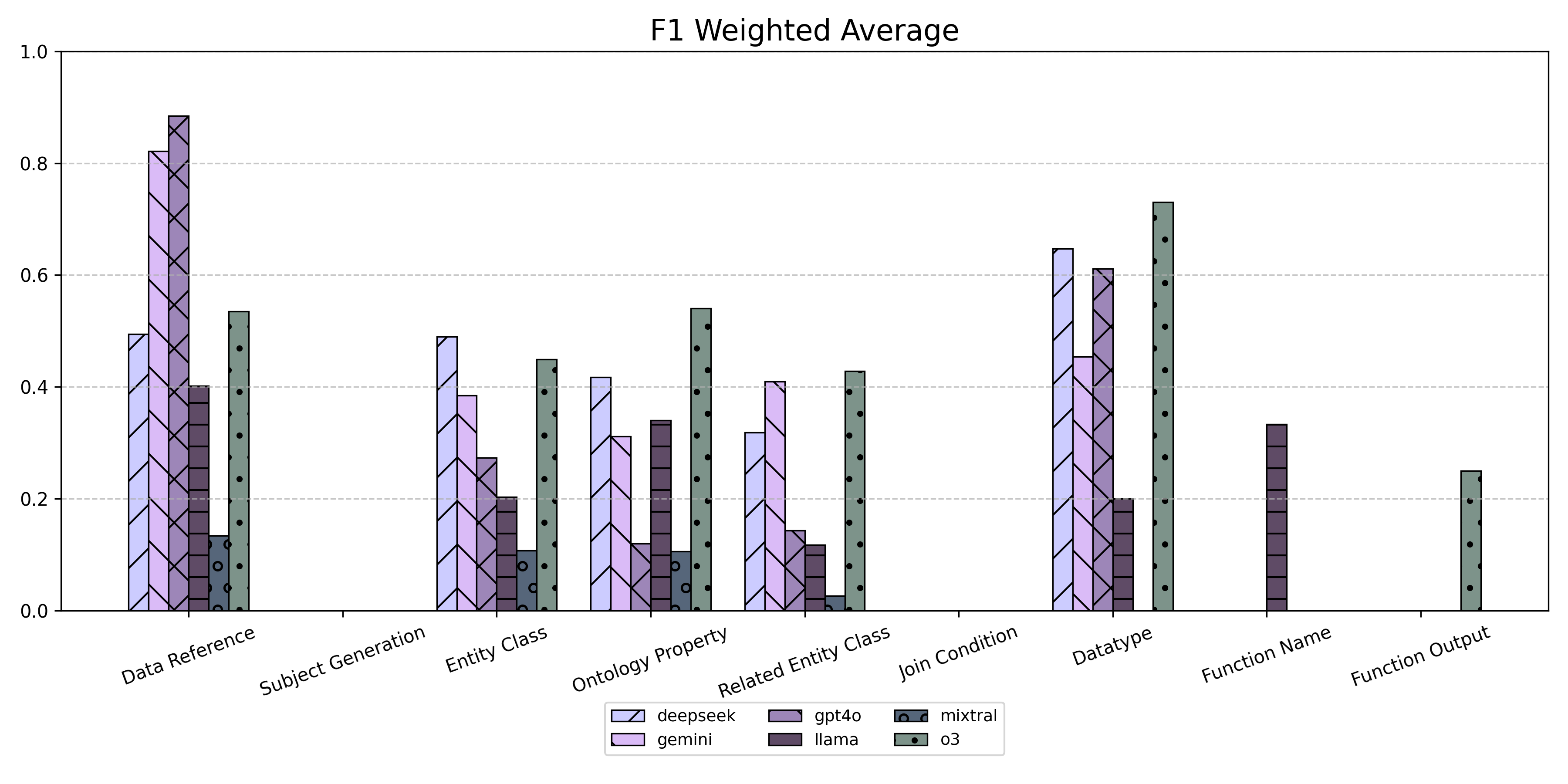}
    \caption{Post-processed F-score of the different tasks and LLMs.}
    \label{fig:results3_auto}
    \end{subfigure}
    \caption{Comparison of the F-score between the three different evaluations in the Scenario 3.}
    \label{fig:f1_sce3}
\end{figure}

F-score evaluations (Figure~\ref{fig:f1_sce3}) reinforce this picture. Manual annotations yield higher scores, especially in concept-level tasks, but even the best-performing models rarely exceed 0.55. Raw evaluations, however, show a marked underestimation of performance due to unstructured outputs, inconsistencies, and hallucinated content. For instance, \textit{Subject Generation} frequently scores near zero in the raw setting, despite partially correct structures. \textit{Join Condition} and \textit{Function Output} also suffer heavily, with most raw scores below 0.10.

Post-processing yields moderate improvements. For example, GPT4o’s score in \textit{Datatype} increases from 0.14 (raw) to 0.31, and similar trends are observed in \textit{Function Name} and \textit{Subject Generation}. This confirms that while relevant information is often present in LLM outputs, superficial noise prevents accurate scoring, and lightweight cleaning helps restore structure. Nonetheless, the gap with manual evaluation remains considerable, especially in tasks requiring symbolic alignment or reasoning.

In Scenario 3, we also analyse the effect of different prompting strategies for DeepSeek. We select DeepSeek as it reports the best performance in the most scenarios. Table \ref{tab:scenario3-results} shows that, while there are noticeable differences across strategies for tasks such as Data Reference identification, Ontology Property and Entity Class recognition, or Subject Generation, all configurations obtain an F-score of 0 for join-related and function-related tasks. These results highlight that, at least for DeepSeek, changing the prompt (zero-shot, one-shot, or few-shot with examples) helps to a certain extent on ``local'' tasks that only require selecting classes, properties, or data references, but it does not suffice to trigger more complex reasoning steps such as specifying joins or data transformation functions. This suggests that prompt engineering helps to obtain better results but alone is not enough to build robust mappings for the more compositional aspects of KG Construction.

\begin{table}[t]
    \centering
    \caption{F-score per task for different prompting strategies with DeepSeek in Scenario~3. Join and function-related tasks are not reported, as all strategies obtain an F-score of 0 on these dimensions.}
    \label{tab:scenario3-deepseek}
    \footnotesize
    \resizebox{\textwidth}{!}{%
    \begin{tabular}{lcccccc}
        \toprule
        \textbf{Method} & \textbf{Data Ref.} & \textbf{Ont. Prop.} & \textbf{Ent. Class} &
        \textbf{Rel. Ent.} & \textbf{Subj. Gen.} & \textbf{Datatype} \\
        \midrule
        0-shot & 0.3750 & 0.3900 & 0.5009 & 0.3852 & 0.0000 & 0.2308 \\
        1-shot & 0.875 & 0.4167 & 0.3553 & 0.2361 & 0.4444 & 0.3333 \\
        few-shot & 0.8636 & 0.3239 & 0.3900 & 0.2222 & 0.7101 & 0.0000 \\
        
        \bottomrule
    \end{tabular}%
    }
\end{table}

In summary, Scenario 3 confirms that schema-distant mappings remain largely out of reach for current general-purpose LLMs. While postprocessing helps mitigate output noise, it does not bridge the semantic gap. These findings underscore the need for hybrid approaches—combining LLMs with ontology-aware reasoning components, intermediate representations, or human-in-the-loop validation—to achieve reliable results in such challenging contexts.

\section{Discussion and Lessons Learned}
\label{sec:discussion}

The BLINKG benchmark reveals both the potential and the limitations of current LLMs when applied to knowledge graph construction through ontology-based mapping generation. Although LLMs demonstrate encouraging results in controlled scenarios, several critical issues emerge when tasks become more complex or less directly aligned with the training distribution of the models. Below, we summarize the main findings and observations:

\begin{itemize}

\item LLMs perform well in simple scenarios, especially when the input schema and ontology share lexical or structural similarities. However, they tend to hallucinate entities or relationships and produce incorrect outputs as the complexity of the mappings increases, particularly in schema-distant cases.
\item Core identification tasks (class, property, related entity) are handled reasonably well, showing low error rates and good semantic alignment with manual annotations. These tasks benefit from contextual cues and general language modeling capabilities.
\item	More technically grounded tasks, such as joins, datatype assignment, or language tagging, remain challenging, with higher error rates and inconsistencies. These tasks often require strict syntax, explicit constraints, or domain knowledge that LLMs do not reliably internalize.
\item The openness of the prompt and input representation has a significant impact on the outcome. Allowing too much flexibility often leads to overly verbose outputs, fabricated mappings, or unintended generalizations. More constrained and structured prompting will tend to yield better results. 
\item Controlled vocabulary linking is an area where LLMs show unexpected strength. Despite the complexity of these tasks for human annotators, due to the need for specialized knowledge, LLMs are often able to identify the correct concept within a taxonomy when enough context is provided.
\item	Evaluation metrics are critical for understanding model behavior. General-purpose metrics are useful but insufficient to capture performance nuances across tasks. We observe that task-specific evaluation measures are likely needed to fairly assess LLM performance, especially for cases involving structure, reasoning, or semantic correctness.
\item A key limitation of current LLM-based benchmarks, including BLINKG, is the potential impact of memorization and data leakage, since some of the ontologies and datasets we use (i.e., GTFS, ePO) are available on the Web and may partially overlap with the models’ training corpora. Although our results already reveal clear weaknesses in the models’ ability to construct semantic mappings, we acknowledge that a more rigorous disentanglement of genuine reasoning from memorised artefacts is needed~\cite{yin-etal-2023-large}. In future iterations of BLINKG should evolve towards the systematic use of synthetic ontologies and input data generated through controlled procedures (i.e. ontology and schema generators), so that the benchmark can better isolate the underlying capabilities of LLMs in KG construction.
\item General-purpose LLMs are unlikely to produce valid solutions in highly complex semantic environments. In these cases, hybrid approaches that combine LLM capabilities with symbolic reasoning, such as OWL-based inference or constraint validation, are essential. Techniques like OWL2RML\footnote{\url{https://github.com/citiususc/owl2yarrrml}} may offer a middle ground between full automation and expert-driven mappings.
\item Mappings should be characterized not only by task type or schema distance, but also by structural features that may impact their generation, such as star-shaped mappings~\cite{iglesias2025empowering}. Capturing these properties can help isolate factors that affect LLM performance and guide more fine-grained benchmarking.  

\item LLMs may be valuable for generating initial mapping drafts, but the involvement of a human expert remains essential. A human-in-the-loop approach is crucial for validating mappings, correcting hallucinations, and ensuring that the resulting knowledge graphs are semantically and structurally sound.
\item Conducting reproducible and in-depth research with proprietary LLMs poses significant challenges. In our case, some of the models initially used in the evaluation, such as OpenAI’s o1~\cite{openai2024o1} and Google’s Gemini 2.0 Pro, were removed or restricted on their respective platforms during the course of the evaluation. This forced us to adapt the benchmark execution and repeat the entire evaluation process with the latest models, highlighting the fragility of relying on closed, commercial systems for scientific experimentation.

\end{itemize}

These observations confirm that, while LLMs represent a promising tool for supporting semantic mapping tasks, their use must be carefully framed within controlled and assisted workflows. BLINKG helps identify where LLMs succeed or fail, and also reveals the importance of hybrid strategies, structured prompting, and task-specific evaluation. As the field advances, we expect that more refined methodologies (combining statistical models with symbolic reasoning and expert oversight) will be key to unlocking the full potential of LLMs in knowledge graph construction.

\section{Conclusion and Future Work}
\label{sec:conclusion}
In this paper, we presented BLINKG, a benchmark specifically designed to evaluate the capabilities of Large Language Models (LLMs) in generating semantic mappings between heterogeneous data sources and ontology terms. Unlike previous efforts, which often focus on end-to-end pipelines or syntactic correctness, BLINKG defines a set of explicit, traceable tasks and provides reusable resources including gold standards, metrics, and diverse scenarios to enable systematic, comparable evaluation across a wide range of conditions. Our experimental results demonstrate that current LLMs are already able to produce promising mappings in well-aligned and controlled scenarios. However, we observe significant limitations when models are faced with schema-distant data, implicit semantics, or complex transformations. In these cases, performance drops notably, particularly in tasks such as conditional joins or function generation. These findings reinforce the need for structured benchmarks like BLINKG to better understand the strengths and weaknesses of LLMs in the context of knowledge graph construction.

Despite its contributions, BLINKG also reveals areas for future improvement. Current evaluation procedures still rely partly on manual inspection or simplified metrics, and handling semantic equivalence beyond syntactic matches remains a challenge. Moreover, the benchmark currently focuses on a core set of tasks; more advanced mapping requirements such as multi-source fusion, list generation, or statement reification are intentionally left for future versions. Finally, as new declarative languages and hybrid AI systems emerge, BLINKG can evolve to include additional modules that evaluate reasoning, transformation chaining, or explainability.


\bibliography{biblio}
\newpage
\appendix
\section{Detailed evaluations of the different scenarios}
\label{appendix1}

The Mean Absolute Error of F-score between expert evaluation and Post-processed in the Scenario 1 are Deepseek: 0.106, Gemini: 0.166, GPT4o: 0.133, o3: 0.169, Llama: 0.123 and Mixtral: 0.236.

\begin{table*}[ht]
\centering
\resizebox{\textwidth}{!}{%
\begin{tabular}{lccc ccc ccc ccc ccc ccc}
\toprule
\textbf{} & \multicolumn{3}{c}{\textbf{deepseek}} & \multicolumn{3}{c}{\textbf{gemini}} & \multicolumn{3}{c}{\textbf{gpt4o}} & \multicolumn{3}{c}{\textbf{o3}} & \multicolumn{3}{c}{\textbf{llama}} & \multicolumn{3}{c}{\textbf{mixtral}} \\
 & Manual & Raw & Post & Manual & Raw & Post & Manual & Raw & Post & Manual & Raw & Post & Manual & Raw & Post & Manual & Raw & Post \\
\midrule
\multicolumn{19}{c}{\textit{Data Reference}} \\
P & 0.84 & 0.88 & 0.88 & 0.92 & 1.00 & 1.00 & 0.81 & 0.75 & 0.75 & 0.76 & 1.00 & 0.88 & 0.84 & 0.88 & 0.63 & 0.80 & 0.63 & 1.00 \\
R & 0.82 & 0.69 & 0.69 & 0.87 & 0.82 & 0.83 & 0.80 & 0.65 & 0.65 & 0.94 & 0.58 & 0.59 & 0.82 & 0.83 & 0.83 & 0.82 & 0.63 & 0.63 \\
F1 & 0.83 & 0.77 & 0.77 & 0.89 & 0.86 & 0.91 & 0.80 & 0.67 & 0.70 & 0.84 & 0.72 & 0.71 & 0.83 & 0.85 & 0.71 & 0.81 & 0.63 & 0.77 \\
\multicolumn{19}{c}{\textit{Ontology Property}} \\
P & 0.94 & 1.00 & 1.00 & 1.00 & 1.00 & 1.00 & 0.94 & 1.00 & 1.00 & 0.82 & 1.00 & 1.00 & 0.90 & 1.00 & 1.00 & 0.81 & 1.00 & 1.00 \\
R & 0.91 & 0.97 & 0.96 & 0.95 & 1.00 & 1.00 & 0.90 & 1.00 & 0.98 & 1.00 & 0.62 & 0.58 & 0.86 & 0.88 & 0.91 & 0.87 & 0.82 & 0.91 \\
F1 & 0.92 & 0.98 & 0.98 & 0.97 & 1.00 & 1.00 & 0.92 & 1.00 & 0.99 & 0.90 & 0.76 & 0.74 & 0.88 & 0.93 & 0.95 & 0.84 & 0.89 & 0.95 \\
\multicolumn{19}{c}{\textit{Entity Class}} \\
P & 0.94 & 1.00 & 1.00 & 1.00 & 1.00 & 1.00 & 0.92 & 1.00 & 1.00 & 0.82 & 1.00 & 1.00 & 0.93 & 1.00 & 1.00 & 0.86 & 1.00 & 1.00 \\
R & 0.91 & 0.98 & 0.98 & 0.95 & 1.00 & 1.00 & 0.90 & 0.99 & 0.99 & 1.00 & 0.93 & 0.92 & 0.89 & 0.99 & 0.99 & 0.90 & 1.00 & 1.00 \\
F1 & 0.92 & 0.99 & 0.99 & 0.97 & 1.00 & 1.00 & 0.91 & 0.99 & 0.99 & 0.90 & 0.96 & 0.96 & 0.91 & 1.00 & 1.00 & 0.88 & 1.00 & 1.00 \\
\multicolumn{19}{c}{\textit{Related Entity Class}} \\
P & 0.80 & 1.00 & 1.00 & 1.00 & 1.00 & 1.00 & 0.87 & 1.00 & 1.00 & 1.00 & 1.00 & 1.00 & 0.78 & 1.00 & 1.00 & 0.87 & 1.00 & 1.00 \\
R & 1.00 & 0.94 & 0.94 & 1.00 & 1.00 & 1.00 & 1.00 & 1.00 & 1.00 & 1.00 & 1.00 & 0.79 & 1.00 & 1.00 & 1.00 & 0.75 & 0.92 & 0.92 \\
F1 & 0.89 & 0.96 & 0.97 & 1.00 & 1.00 & 1.00 & 0.93 & 1.00 & 1.00 & 1.00 & 1.00 & 0.88 & 0.87 & 1.00 & 1.00 & 0.79 & 0.95 & 0.96 \\
\multicolumn{19}{c}{\textit{Subject Generation}} \\
P & 0.96 & 0.75 & 0.75 & 1.00 & 1.00 & 1.00 & 0.90 & 1.00 & 1.00 & 0.82 & 0.75 & 0.75 & 0.93 & 1.00 & 1.00 & 0.50 & 0.50 & 0.50 \\
R & 0.91 & 0.47 & 0.47 & 0.94 & 0.58 & 0.58 & 0.88 & 0.77 & 0.77 & 1.00 & 0.65 & 0.68 & 0.89 & 0.87 & 0.87 & 0.49 & 0.25 & 0.25 \\
F1 & 0.93 & 0.57 & 0.58 & 0.97 & 0.68 & 0.73 & 0.89 & 0.84 & 0.87 & 0.90 & 0.69 & 0.72 & 0.91 & 0.92 & 0.93 & 0.50 & 0.30 & 0.34 \\
\multicolumn{19}{c}{\textit{Join Condition}} \\
P & 0.40 & 0.50 & 0.50 & 0.92 & 0.25 & 0.50 & 0.32 & 0.25 & 0.25 & 0.79 & 0.50 & 0.75 & 0.15 & 0.00 & 0.00 & 0.33 & 0.00 & 0.00 \\
R & 0.58 & 0.50 & 0.50 & 0.92 & 0.17 & 0.42 & 0.67 & 0.08 & 0.08 & 1.00 & 0.50 & 0.63 & 0.33 & 0.00 & 0.00 & 0.33 & 0.00 & 0.00 \\
F1 & 0.47 & 0.50 & 0.50 & 0.92 & 0.20 & 0.45 & 0.43 & 0.13 & 0.12 & 0.84 & 0.50 & 0.68 & 0.21 & 0.00 & 0.00 & 0.32 & 0.00 & 0.00 \\
\multicolumn{19}{c}{\textit{Datatype}} \\
P & 0.92 & 1.00 & 1.00 & 1.00 & 1.00 & 1.00 & 0.89 & 1.00 & 1.00 & 0.77 & 1.00 & 1.00 & 0.82 & 1.00 & 1.00 & 0.71 & 1.00 & 1.00 \\
R & 0.83 & 0.96 & 0.97 & 0.92 & 1.00 & 1.00 & 0.83 & 0.98 & 1.00 & 1.00 & 0.59 & 0.65 & 0.83 & 0.91 & 0.92 & 0.72 & 0.91 & 0.85 \\
F1 & 0.87 & 0.98 & 0.98 & 0.96 & 1.00 & 1.00 & 0.86 & 0.99 & 1.00 & 0.87 & 0.71 & 0.79 & 0.82 & 0.95 & 0.96 & 0.71 & 0.94 & 0.92 \\
\multicolumn{19}{c}{\textit{Language Annotations}} \\
P & 1.00 & 1.00 & 1.00 & 1.00 & 0.50 & 0.50 & 0.81 & 1.00 & 1.00 & 0.87 & 0.50 & 0.50 & 0.58 & 0.50 & 0.50 & 0.83 & 0.00 & 0.00 \\
R & 1.00 & 0.83 & 0.83 & 1.00 & 0.50 & 0.50 & 0.78 & 1.00 & 1.00 & 1.00 & 0.43 & 0.43 & 1.00 & 0.50 & 0.50 & 0.78 & 0.00 & 0.00 \\
F1 & 1.00 & 0.90 & 0.91 & 1.00 & 0.50 & 0.50 & 0.73 & 1.00 & 1.00 & 0.92 & 0.46 & 0.46 & 0.70 & 0.50 & 0.50 & 0.76 & 0.00 & 0.00 \\
\bottomrule
\end{tabular}
}
\caption{Evaluation metrics (Precision, Recall, F1-score) for Scenario 1 across all models and evaluation types.}
\label{tab:scenario1-results}
\end{table*}

\newpage
The Mean Absolute Error of F-score between expert evaluation and Post-processed in the Scenario 2 are Deepseek: 0.16, Gemini: 0.22, GPT4o: 0.13, o3: 0.21, Llama: 0.21 and Mixtral: 0.22.

\begin{table*}[!h]
\centering
\resizebox{\textwidth}{!}{%
\begin{tabular}{lccc ccc ccc ccc ccc ccc}
\toprule
\textbf{} & \multicolumn{3}{c}{\textbf{deepseek}} & \multicolumn{3}{c}{\textbf{gemini}} & \multicolumn{3}{c}{\textbf{gpt4o}} & \multicolumn{3}{c}{\textbf{o3}} & \multicolumn{3}{c}{\textbf{llama}} & \multicolumn{3}{c}{\textbf{mixtral}} \\
 & Manual & Raw & Post & Manual & Raw & Post & Manual & Raw & Post & Manual & Raw & Post & Manual & Raw & Post & Manual & Raw & Post \\
\midrule
\multicolumn{19}{c}{\textit{Data Reference}} \\
P & 0.85 & 1.00 & 1.00 & 0.84 & 1.00 & 1.00 & 0.85 & 1.00 & 1.00 & 0.84 & 1.00 & 1.00 & 0.88 & 1.00 & 1.00 & 0.90 & 1.00 & 1.00 \\
R & 0.96 & 0.80 & 0.91 & 1.00 & 0.72 & 0.95 & 1.00 & 0.86 & 0.95 & 0.94 & 0.63 & 0.79 & 0.96 & 0.76 & 0.96 & 0.94 & 0.82 & 0.94 \\
F1 & 0.90 & 0.88 & 0.95 & 0.91 & 0.82 & 0.97 & 0.92 & 0.92 & 0.97 & 0.92 & 0.74 & 0.86 & 0.92 & 0.83 & 0.98 & 0.92 & 0.88 & 0.96 \\
\multicolumn{19}{c}{\textit{Ontology Property}} \\
P & 0.98 & 1.00 & 1.00 & 0.99 & 1.00 & 1.00 & 0.99 & 1.00 & 1.00 & 1.00 & 1.00 & 1.00 & 1.00 & 1.00 & 1.00 & 0.98 & 1.00 & 1.00 \\
R & 0.94 & 0.92 & 0.98 & 1.00 & 0.84 & 0.98 & 0.99 & 0.93 & 0.99 & 1.00 & 0.84 & 1.00 & 0.93 & 0.83 & 1.00 & 0.90 & 0.87 & 0.95 \\
F1 & 0.96 & 0.95 & 0.99 & 1.00 & 0.91 & 0.99 & 0.99 & 0.96 & 0.99 & 1.00 & 0.91 & 1.00 & 0.96 & 0.89 & 1.00 & 0.94 & 0.92 & 0.98 \\
\multicolumn{19}{c}{\textit{Entity Class}} \\
P & 0.93 & 0.83 & 1.00 & 0.95 & 1.00 & 1.00 & 0.99 & 1.00 & 1.00 & 0.95 & 1.00 & 1.00 & 0.78 & 1.00 & 1.00 & 0.82 & 1.00 & 1.00 \\
R & 0.90 & 0.57 & 0.65 & 0.95 & 0.82 & 0.95 & 0.99 & 0.86 & 0.94 & 0.95 & 0.73 & 0.90 & 0.73 & 0.66 & 0.79 & 0.74 & 0.71 & 0.77 \\
F1 & 0.91 & 0.62 & 0.73 & 0.95 & 0.89 & 0.97 & 0.99 & 0.92 & 0.96 & 0.95 & 0.82 & 0.93 & 0.75 & 0.72 & 0.83 & 0.78 & 0.78 & 0.82 \\
\multicolumn{19}{c}{\textit{Related Entity Class}} \\
P & 0.84 & 0.67 & 0.83 & 0.98 & 0.83 & 1.00 & 1.00 & 0.83 & 1.00 & 1.00 & 0.33 & 0.50 & 0.12 & 0.50 & 0.67 & 0.41 & 0.83 & 1.00 \\
R & 0.73 & 0.38 & 0.54 & 1.00 & 0.81 & 0.97 & 0.92 & 0.79 & 0.98 & 0.99 & 0.32 & 0.50 & 0.14 & 0.38 & 0.59 & 0.35 & 0.66 & 0.82 \\
F1 & 0.77 & 0.47 & 0.63 & 0.99 & 0.82 & 0.98 & 0.96 & 0.81 & 0.99 & 0.99 & 0.32 & 0.50 & 0.13 & 0.43 & 0.62 & 0.37 & 0.72 & 0.88 \\
\multicolumn{19}{c}{\textit{Subject Generation}} \\
P & 0.95 & 1.00 & 1.00 & 0.99 & 0.83 & 0.83 & 0.97 & 1.00 & 1.00 & 0.95 & 0.17 & 0.83 & 0.82 & 0.33 & 0.33 & 0.45 & 0.17 & 0.17 \\
R & 0.53 & 0.37 & 0.68 & 1.00 & 0.23 & 0.30 & 0.93 & 0.51 & 0.63 & 0.95 & 0.03 & 0.59 & 0.76 & 0.13 & 0.22 & 0.34 & 0.04 & 0.06 \\
F1 & 0.68 & 0.50 & 0.75 & 0.99 & 0.34 & 0.42 & 0.95 & 0.62 & 0.71 & 0.95 & 0.05 & 0.67 & 0.79 & 0.18 & 0.27 & 0.39 & 0.07 & 0.08 \\
\multicolumn{19}{c}{\textit{Join Condition}} \\
P & 0.23 & 0.50 & 0.67 & 0.61 & 0.67 & 0.67 & 0.36 & 0.83 & 0.83 & 0.38 & 0.83 & 1.00 & 0.03 & 0.00 & 0.00 & 0.05 & 0.33 & 0.33 \\
R & 0.53 & 0.33 & 0.50 & 0.73 & 0.57 & 0.59 & 0.70 & 0.53 & 0.56 & 0.80 & 0.64 & 0.79 & 0.13 & 0.00 & 0.00 & 0.10 & 0.14 & 0.14 \\
F1 & 0.32 & 0.40 & 0.55 & 0.66 & 0.61 & 0.62 & 0.45 & 0.63 & 0.65 & 0.50 & 0.70 & 0.85 & 0.05 & 0.00 & 0.00 & 0.06 & 0.19 & 0.19 \\
\multicolumn{19}{c}{\textit{Datatype}} \\
P & 0.79 & 1.00 & 1.00 & 0.90 & 1.00 & 1.00 & 0.89 & 1.00 & 1.00 & 0.76 & 1.00 & 1.00 & 0.74 & 0.83 & 0.83 & 0.84 & 1.00 & 1.00 \\
R & 0.77 & 0.67 & 0.70 & 0.90 & 0.77 & 0.82 & 0.91 & 0.74 & 0.79 & 0.77 & 0.40 & 0.52 & 0.69 & 0.63 & 0.64 & 0.72 & 0.59 & 0.70 \\
F1 & 0.78 & 0.79 & 0.81 & 0.90 & 0.86 & 0.89 & 0.90 & 0.81 & 0.86 & 0.77 & 0.56 & 0.68 & 0.71 & 0.69 & 0.72 & 0.77 & 0.69 & 0.80 \\
\multicolumn{19}{c}{\textit{Function Name}} \\
P & 1.00 & 0.00 & 0.20 & 0.98 & 0.00 & 0.20 & 0.86 & 0.00 & 0.20 & 0.71 & 0.00 & 0.20 & 1.00 & 0.00 & 0.20 & 0.80 & 0.00 & 0.00 \\
R & 0.81 & 0.00 & 0.20 & 0.76 & 0.00 & 0.20 & 0.67 & 0.00 & 0.20 & 0.78 & 0.00 & 0.20 & 0.76 & 0.00 & 0.20 & 0.56 & 0.00 & 0.00 \\
F1 & 0.89 & 0.00 & 0.20 & 0.86 & 0.00 & 0.20 & 0.75 & 0.00 & 0.20 & 0.74 & 0.00 & 0.20 & 0.86 & 0.00 & 0.20 & 0.64 & 0.00 & 0.00 \\
\multicolumn{19}{c}{\textit{Function Output}} \\
P & 0.93 & 0.67 & 1.00 & 0.33 & 0.83 & 1.00 & 0.75 & 0.33 & 1.00 & 0.75 & 0.33 & 1.00 & 0.76 & 0.33 & 0.80 & 0.30 & 0.17 & 1.00 \\
R & 0.81 & 0.27 & 0.79 & 0.25 & 0.43 & 0.80 & 0.65 & 0.18 & 0.63 & 0.70 & 0.08 & 0.58 & 0.67 & 0.19 & 0.68 & 0.24 & 0.06 & 0.35 \\
F1 & 0.86 & 0.34 & 0.86 & 0.29 & 0.52 & 0.86 & 0.69 & 0.21 & 0.73 & 0.72 & 0.13 & 0.67 & 0.71 & 0.24 & 0.73 & 0.26 & 0.08 & 0.49 \\
\bottomrule
\end{tabular}
}
\caption{Evaluation metrics (Precision, Recall, F1-score) for Scenario 2 across all models and evaluation types.}
\label{tab:scenario2-results}
\end{table*}

\newpage
The Mean Absolute Error of F-score between expert evaluation and Post-processed in the Scenario 3 are 	Deepseek: 0.42, Gemini: 0.38, GPT4o: 0.35, o3: 0.32, LLaMa: 0.35, Mixtral: 0.34
\begin{table*}[!h]
\centering
\resizebox{\textwidth}{!}{%
\begin{tabular}{lccc ccc ccc ccc ccc ccc}
\toprule
\textbf{} & \multicolumn{3}{c}{\textbf{deepseek}} & \multicolumn{3}{c}{\textbf{gemini}} & \multicolumn{3}{c}{\textbf{gpt4o}} & \multicolumn{3}{c}{\textbf{o3}} & \multicolumn{3}{c}{\textbf{llama}} & \multicolumn{3}{c}{\textbf{mixtral}} \\
 & Manual & Raw & Post & Manual & Raw & Post & Manual & Raw & Post & Manual & Raw & Post & Manual & Raw & Post & Manual & Raw & Post \\
\midrule
\multicolumn{19}{c}{\textit{Data Reference}} \\
P & 0.05 & 1.00 & 1.00 & 0.07 & 1.00 & 1.00 & 0.08 & 1.00 & 1.00 & 0.02 & 1.00 & 1.00 & 0.02 & 1.00 & 1.00 & 0.00 & 0.50 & 0.50 \\
R & 0.05 & 0.24 & 0.33 & 0.10 & 0.64 & 0.70 & 0.06 & 0.84 & 0.79 & 0.03 & 0.24 & 0.40 & 0.03 & 0.31 & 0.29 & 0.00 & 0.17 & 0.08 \\
F1 & 0.05 & 0.39 & 0.49 & 0.09 & 0.78 & 0.82 & 0.07 & 0.91 & 0.88 & 0.02 & 0.38 & 0.54 & 0.02 & 0.42 & 0.40 & 0.00 & 0.25 & 0.13 \\
\multicolumn{19}{c}{\textit{Ontology Property}} \\
P & 0.42 & 0.50 & 1.00 & 0.39 & 0.50 & 1.00 & 0.34 & 0.50 & 1.00 & 0.53 & 1.00 & 1.00 & 0.21 & 0.00 & 0.50 & 0.15 & 0.00 & 0.00 \\
R & 0.59 & 0.15 & 0.51 & 0.53 & 0.06 & 0.31 & 0.22 & 0.04 & 0.56 & 0.58 & 0.17 & 0.60 & 0.31 & 0.00 & 0.13 & 0.15 & 0.00 & 0.00 \\
F1 & 0.49 & 0.23 & 0.65 & 0.44 & 0.11 & 0.45 & 0.26 & 0.07 & 0.61 & 0.53 & 0.29 & 0.73 & 0.25 & 0.00 & 0.20 & 0.15 & 0.00 & 0.00 \\
\multicolumn{19}{c}{\textit{Entity Class}} \\
P & 0.40 & 1.00 & 1.00 & 0.39 & 1.00 & 1.00 & 0.34 & 1.00 & 1.00 & 0.51 & 1.00 & 1.00 & 0.16 & 1.00 & 1.00 & 0.05 & 1.00 & 1.00 \\
R & 0.56 & 0.33 & 0.32 & 0.54 & 0.14 & 0.25 & 0.22 & 0.09 & 0.16 & 0.56 & 0.18 & 0.30 & 0.23 & 0.10 & 0.11 & 0.05 & 0.06 & 0.06 \\
F1 & 0.47 & 0.50 & 0.49 & 0.45 & 0.25 & 0.38 & 0.26 & 0.17 & 0.27 & 0.54 & 0.30 & 0.45 & 0.18 & 0.18 & 0.20 & 0.05 & 0.11 & 0.11 \\
\multicolumn{19}{c}{\textit{Related Entity Class}} \\
P & 0.52 & 1.00 & 1.00 & 0.51 & 1.00 & 1.00 & 0.31 & 0.50 & 0.50 & 0.60 & 1.00 & 1.00 & 0.08 & 0.50 & 0.50 & 0.07 & 0.00 & 0.50 \\
R & 0.54 & 0.24 & 0.19 & 0.56 & 0.16 & 0.27 & 0.19 & 0.11 & 0.08 & 0.67 & 0.19 & 0.30 & 0.13 & 0.05 & 0.07 & 0.10 & 0.00 & 0.01 \\
F1 & 0.53 & 0.39 & 0.32 & 0.53 & 0.27 & 0.41 & 0.23 & 0.18 & 0.14 & 0.63 & 0.31 & 0.43 & 0.09 & 0.09 & 0.12 & 0.08 & 0.00 & 0.03 \\
\multicolumn{19}{c}{\textit{Subject Generation}} \\
P & 0.00 & 0.00 & 0.00 & 0.07 & 0.00 & 0.00 & 0.05 & 0.00 & 0.00 & 0.00 & 0.00 & 0.00 & 0.00 & 0.00 & 0.00 & 0.00 & 0.00 & 0.00 \\
R & 0.00 & 0.00 & 0.00 & 0.08 & 0.00 & 0.00 & 0.03 & 0.00 & 0.00 & 0.00 & 0.00 & 0.00 & 0.00 & 0.00 & 0.00 & 0.00 & 0.00 & 0.00 \\
F1 & 0.00 & 0.00 & 0.00 & 0.07 & 0.00 & 0.00 & 0.03 & 0.00 & 0.00 & 0.00 & 0.00 & 0.00 & 0.00 & 0.00 & 0.00 & 0.00 & 0.00 & 0.00 \\
\multicolumn{19}{c}{\textit{Join Condition}} \\
P & 0.00 & 0.00 & 0.00 & 0.04 & 0.00 & 0.00 & 0.00 & 0.00 & 0.00 & 0.00 & 0.00 & 0.00 & 0.00 & 0.00 & 0.00 & 0.00 & 0.00 & 0.00 \\
R & 0.00 & 0.00 & 0.00 & 0.07 & 0.00 & 0.00 & 0.00 & 0.00 & 0.00 & 0.00 & 0.00 & 0.00 & 0.00 & 0.00 & 0.00 & 0.00 & 0.00 & 0.00 \\
F1 & 0.00 & 0.00 & 0.00 & 0.05 & 0.00 & 0.00 & 0.00 & 0.00 & 0.00 & 0.00 & 0.00 & 0.00 & 0.00 & 0.00 & 0.00 & 0.00 & 0.00 & 0.00 \\
\multicolumn{19}{c}{\textit{Datatype}} \\
P & 0.22 & 1.00 & 1.00 & 0.25 & 1.00 & 1.00 & 0.89 & 0.50 & 0.50 & 0.44 & 1.00 & 1.00 & 0.09 & 1.00 & 1.00 & 0.10 & 0.50 & 0.50 \\
R & 0.48 & 0.25 & 0.27 & 0.67 & 0.12 & 0.18 & 0.91 & 0.07 & 0.07 & 0.67 & 0.19 & 0.38 & 0.10 & 0.18 & 0.20 & 0.10 & 0.01 & 0.06 \\
F1 & 0.30 & 0.39 & 0.42 & 0.36 & 0.21 & 0.31 & 0.90 & 0.12 & 0.12 & 0.52 & 0.32 & 0.54 & 0.09 & 0.30 & 0.34 & 0.10 & 0.02 & 0.11 \\
\multicolumn{19}{c}{\textit{Function Name}} \\
P & 0.78 & 0.00 & 0.00 & 0.62 & 0.00 & 0.00 & 0.47 & 0.00 & 0.00 & 0.75 & 0.00 & 0.00 & 0.67 & 0.50 & 0.50 & 0.00 & 0.00 & 0.00 \\
R & 0.92 & 0.00 & 0.00 & 0.83 & 0.00 & 0.00 & 0.42 & 0.00 & 0.00 & 0.75 & 0.00 & 0.00 & 0.42 & 0.08 & 0.25 & 0.00 & 0.00 & 0.00 \\
F1 & 0.84 & 0.00 & 0.00 & 0.71 & 0.00 & 0.00 & 0.44 & 0.00 & 0.00 & 0.75 & 0.00 & 0.00 & 0.44 & 0.14 & 0.33 & 0.00 & 0.00 & 0.00 \\
\multicolumn{19}{c}{\textit{Function Output}} \\
P & 0.00 & 0.00 & 0.00 & 0.00 & 0.00 & 0.00 & 0.00 & 0.00 & 0.00 & 0.00 & 0.00 & 0.50 & 0.00 & 0.00 & 0.00 & 0.00 & 0.00 & 0.00 \\
R & 0.00 & 0.00 & 0.00 & 0.00 & 0.00 & 0.00 & 0.00 & 0.00 & 0.00 & 0.00 & 0.00 & 0.17 & 0.00 & 0.00 & 0.00 & 0.00 & 0.00 & 0.00 \\
F1 & 0.00 & 0.00 & 0.00 & 0.00 & 0.00 & 0.00 & 0.00 & 0.00 & 0.00 & 0.00 & 0.00 & 0.25 & 0.00 & 0.00 & 0.00 & 0.00 & 0.00 & 0.00 \\
\bottomrule
\end{tabular}
}
\caption{Evaluation metrics (Precision, Recall, F1-score) for Scenario 3 across all models and evaluation types.}
\label{tab:scenario3-results}
\end{table*}

\appendix

\end{document}